\documentclass[letterpaper]{article} 
\usepackage[preprint]{aaai2027}
\usepackage[hyphens]{url}  
\usepackage{graphicx} 
\usepackage{natbib}  
\usepackage{caption} 
\usepackage{algorithm}
\usepackage{algorithmic}

\usepackage{newfloat}
\usepackage{listings}

\usepackage{xspace}
\usepackage{amsmath}
\usepackage{amssymb}
\usepackage{multirow}
\usepackage{subcaption}

\DeclareCaptionStyle{ruled}{labelfont=normalfont,labelsep=colon,strut=off} 
\floatstyle{ruled}
\newfloat{listing}{tb}{lst}{}
\floatname{listing}{Listing}

\usepackage{booktabs}

\newcommand{\method}{HiPPT\xspace}
\title{
	\method: Physiology-Grounded Wearable PPG Representation Learning via Multi-Scope Text Alignment
}
\author{
	Jiaqi Wang\textsuperscript{1}\equalcontrib, Wei Chen\textsuperscript{1}\equalcontrib, Xiaoyu Li\textsuperscript{2}, Lexian Xia\textsuperscript{3}, ShuaiLong Tang\textsuperscript{3}, Chenran Zhang\textsuperscript{1}, Donglin Xie\textsuperscript{4}, Guangpu Zhu\textsuperscript{2}, Yelei Li\textsuperscript{2}\corresponding, and Yi Zhou\textsuperscript{1}\corresponding
}
\affiliations{
	\textsuperscript{1}Southeast University \quad
	\textsuperscript{2}OPPO \quad
	\textsuperscript{3}Shenzhen Technology University \quad
	\textsuperscript{4}Peking University\\
	\texttt{larry611@gmail.com, yizhou.szcn@gmail.com}
}

\begin{document}

\maketitle

\begin{abstract}

Reusable wearable PPG representations should capture local pulse morphology and broader physiological regularities. Existing pretraining objectives, however, usually apply supervision at the recording or pair level, overlooking that different concepts become reliable over different scopes of waveform evidence. We propose HiPPT, a physiology-grounded framework that matches each supervision source to its appropriate evidence scope. We construct a PPG-Text corpus of over 4 million wrist-worn recordings from 202K users, pairing recordings with waveform-derived morphology and metadata-derived context descriptions. HiPPT combines record-level morphology alignment, which grounds individual recordings in pulse characteristics, with cross-recording context alignment, which aggregates recordings sharing the same metadata tuple to capture context-consistent regularities. Waveform-guided text modeling further recovers masked contextual concepts from PPG patch representations, providing fine-grained waveform-text supervision. Across six downstream tasks spanning blood pressure, vascular stiffness, physiological attributes, and cross-dataset heart-rate estimation, HiPPT achieves the best result on the primary linear-probing metrics. The aligned space also supports zero-shot profiling of age group, sex, and BMI category, individually and jointly, with performance improving as more recordings are aggregated. These results show that evidence-scope matching yields transferable and semantically accessible wearable PPG representations. Code and pretrained weights will be released.

\end{abstract}


\section{Introduction}
Photoplethysmography (PPG) has become a widely available sensing modality in wrist-worn devices, enabling continuous monitoring of cardiovascular and health-related conditions in daily life~\cite{ppg0,ppg3,ppg5}.
Recent advances in large-scale pretraining have further demonstrated the potential of wearable PPG for learning reusable representations across diverse physiological tasks~\cite{apple0,papagei,saha2025pulse,chen2025gpt}.
However, a wearable PPG recording is an entangled observation that jointly reflects cardiovascular dynamics, transient physiological variation, sensor-skin contact, motion, peripheral perfusion, and other acquisition-related disturbances~\cite{park2022photoplethysmogram}.
The challenge in PPG pretraining is therefore not only to scale data and model capacity, but also to determine which physiological regularities should organize the learned representation.


Existing PPG pretraining methods derive supervision from diverse sources.
Augmentation- and reconstruction-based objectives exploit transformed or incomplete recordings~\cite{narayanswamy2024scaling,chen2025gpt}, while morphology-aware, quality-aware, and subject-aware methods incorporate domain priors or predefined relations across samples~\cite{apple_1,papagei,saha2025pulse,ding2024siamquality}.
Despite their effectiveness, these methods mainly vary the source of supervision while keeping its application scope at the recording or pair level.
Consequently, existing objectives rarely distinguish whether a supervisory concept is reliably observable within one segment or becomes stable only across repeated observations.

Different supervisory concepts, however, become reliable over different scopes of waveform evidence.
Waveform morphology, such as cardiac-cycle duration, systolic and diastolic timing, pulse contour, and peak intervals, is directly observable from an individual PPG recording.
In contrast, physiological regularities associated with demographic, anthropometric, and temporal context may be only weakly and indirectly reflected in any single segment with limited temporal coverage.
Their manifestations can be obscured by motion artifacts, sensor contact, peripheral perfusion, and transient physiological state.
Multiple recordings under comparable contexts provide repeated observations, allowing context-consistent regularities to emerge despite record-specific variation.
This motivates a general principle: the unit over which supervision is applied should match the scope over which its associated waveform evidence becomes reliable.
We leverage hand-crafted waveform features and user metadata along with the acquisition timestamp to implement this principle.

Among possible encodings of such supervision, language is particularly attractive because it allows both sources of supervision to reuse a representation space initialized by a pretrained text encoder~\cite{clip,li2023frozenlanguagemodelhelps,etp,sensorlm}, rather than learning the semantic space entirely from scratch.
Specifically, domain-informed waveform features are converted into record-specific morphology descriptions, while routinely available metadata are serialized into contextual descriptions containing both quantitative values and qualitative descriptors.
Once aligned with text, the PPG representation space can also be queried using textual descriptions.
Based on this formulation, we construct a template-curated PPG-Text pretraining dataset containing over 4M wrist-worn recordings from approximately 202K real-world users.
Another consideration beyond evidence scope is that global alignment may capture coarse sentence-level similarity, without specifying how the constituent concepts should interact with PPG representations.

To address issues mentioned above, we introduce \textbf{\method}, a physiology-grounded PPG representation learning framework with supervision applied at multiple evidence scopes.
Record-Level Morphology Alignment associates each recording with its waveform-derived description, preserving directly observable and record-specific morphological information.
To organize isolated recordings around context-consistent physiological structure, Cross-Recording Context Alignment aggregates recordings sharing the same metadata and aligns the resulting representation with an associated contextual description.
To encourage fine-grained PPG representation learning, Waveform-Guided Text Modeling objective recovers the masked metadata-relevant textual content conditioned jointly on the remaining text and PPG patch representations.

We evaluate \method on six downstream tasks spanning blood-pressure estimation, vascular-function assessment, physiological-attribute prediction, and cross-dataset heart-rate estimation.
\method achieves the best primary linear-probing performance for systolic blood pressure, diastolic blood pressure, and pulse-wave velocity, while transferring effectively to the public PPG-DaLiA dataset~\cite{ppg3}.
Beyond conventional transfer learning, the aligned representation space supports zero-shot user profiling of age group, sex, and BMI category, both individually and jointly.
Zero-shot profiling further improves as more recordings are aggregated, consistent with the proposed role of cross-recording evidence accumulation.
Our main contributions are as follows:
\begin{itemize}
    \item We construct a large-scale PPG-Text corpus containing over 4M wrist-worn recordings from approximately 202K users, with complementary descriptions derived from waveform morphology and routinely available metadata.
    \item We propose \method, a PPG representation learning framework that matches each source of supervision to its appropriate evidence scope by aligning morphology descriptions with individual recordings and contextual descriptions with aggregated recording representations. Waveform-Guided Text Modeling further provides fine-grained patch-level supervision.
    \item Across six downstream tasks and zero-shot user profiling, we demonstrate that \method learns transferable and semantically accessible PPG representations.
\end{itemize}

\begin{figure*}[!t]
    \centering
    \includegraphics[width=0.9\linewidth]{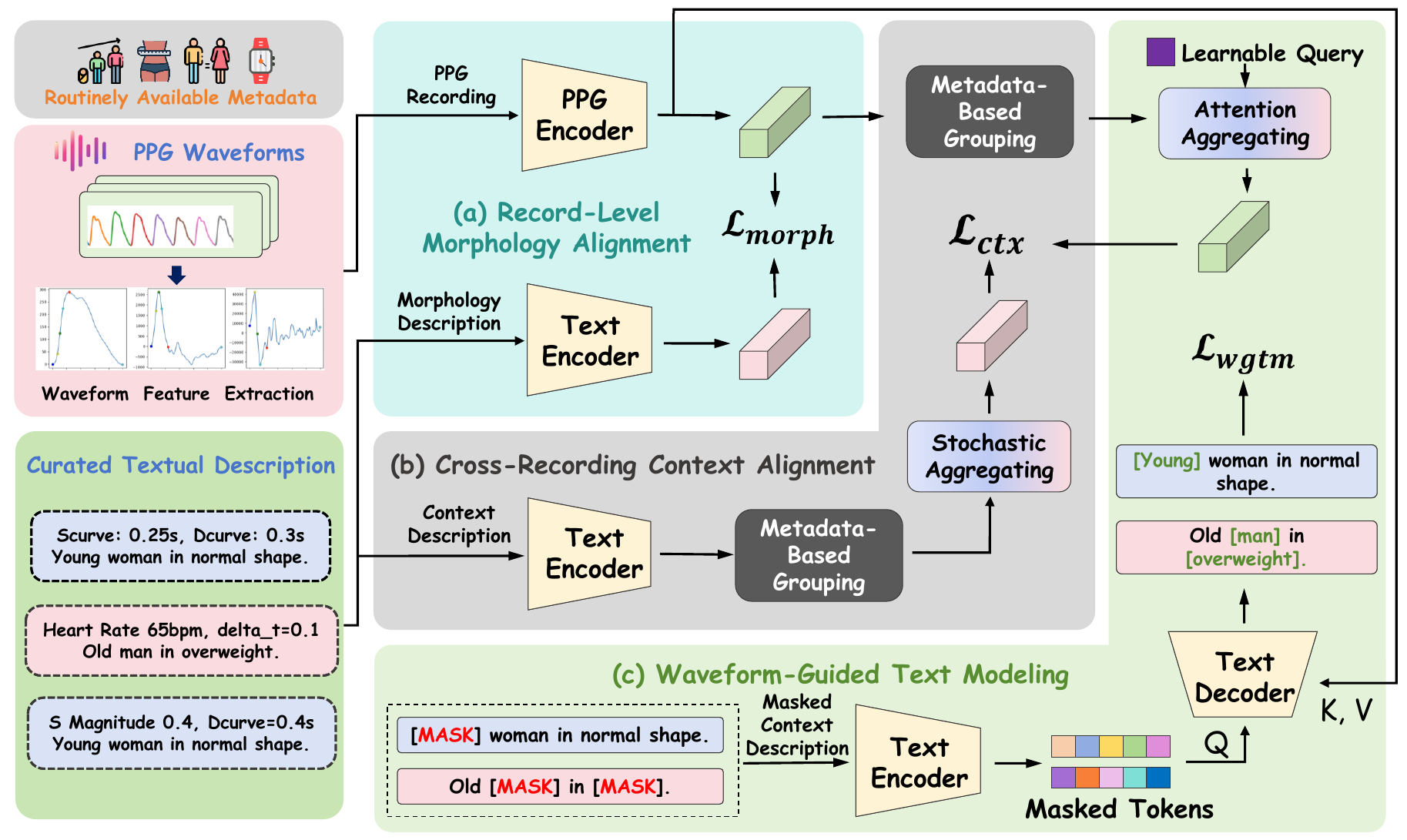}
    \caption{Overview of \method. Record-Level Morphology Alignment grounds each PPG recording in its waveform-derived morphology description. Cross-Recording Context Alignment aggregates a collection of recordings sharing the same metadata tuple and aligns the aggregated representation with the corresponding context description. Waveform-Guided Text Modeling predicts masked contextual concepts conditioned on the remaining text and PPG patch representations, providing fine-grained waveform–text supervision.}
    \label{fig:method:overview}
\end{figure*}

\section{Related Work}
\subsection{PPG Representation Learning}

Self-supervised PPG representation learning has primarily exploited supervision derived from the signal itself or from relations among recordings~\cite{apple0,papagei,chen2025gpt}.
Contrastive methods construct positive pairs using signal augmentations, subject identity, or domain-specific physiological similarity.
Abbaspourazad et al. and Cheng et al.~\cite{apple0,apple_1}, for example, used recordings from the same subject to encourage invariance to inter-subject variation.
TF-C~\cite{zhang2022self} aligned temporal and frequency-domain representations, while PaPaGei~\cite{papagei} and related approaches incorporated waveform morphology, recurring motifs, temporal physiological states, or signal quality into contrastive objectives~\cite{saha2025pulse,ma2024upr,ding2024siamquality}.
Reconstruction, distillation, and generative objectives provide additional routes to learning robust signal structure.
Cross-modal and quality-aware distillation have transferred PPG-derived or signal-quality knowledge across wearable modalities and recording conditions~\cite{abbaspourazad2024wearable,guo2025qualityfm}.
Masked-autoencoding approaches have also been scaled to large wearable datasets~\cite{narayanswamy2024scaling}, while generative pretraining has been explored for modeling physiological time series~\cite{chen2025gpt}.

These studies demonstrate that PPG representations benefit from both waveform-specific priors and carefully designed relations between samples.
Nevertheless, their supervision is generally organized around an individual recording, its augmented views, or pairwise relations such as shared subject identity.
Contextual metadata, when used, are commonly introduced as auxiliary prediction targets or fixed labels~\cite{pulselm,cap}.
Consequently, regularities that recur across multiple recordings under comparable demographic, anthropometric, and temporal contexts have received less explicit treatment as a distinct level of supervision.

\subsection{Biosignal-Language Representation Learning}

Inspired by vision-language pretraining such as CLIP~\cite{clip}, recent work has introduced natural language as semantic supervision for physiological signals.
METS and ETP~\cite{li2023frozenlanguagemodelhelps,etp} aligned ECG signals with clinical reports in a shared embedding space, supporting zero-shot recognition and transferable representations.
GEM~\cite{GEM} further combined ECG time series, waveform images, and textual information within a unified semantic space.
Beyond global signal-text alignment, MELP~\cite{token2rythm} introduced supervision at the token, heartbeat, and rhythm levels to capture the internal temporal hierarchy of ECG.
SensorLM~\cite{sensorlm} developed a captioning pipeline that converts statistical, structural, and semantic characteristics of wearable sensor data into natural-language descriptions at scale.

These advances show that language can connect heterogeneous signal properties through a common representation space.
Existing multi-scale approaches, however, mainly define granularity within an individual recording or its temporal structure.
Unlike clinical ECG, wearable PPG is typically collected without report-like text, making the construction and organization of language supervision a central challenge.
Recent PPG-specific efforts have begun to explore several solutions.
PulseLM~\cite{pulselm} reformulated heterogeneous annotations from public PPG datasets as closed-ended question-answer pairs for multimodal physiological reasoning, whereas CAP~\cite{cap} aligned PPG with patient-level clinical semantics distilled from electronic health records.
In contrast, \method learns from wearable PPG without relying on paired clinical reports.
It constructs complementary descriptions from domain-informed waveform features and contextual metadata, and organizes alignment both at the individual-recording level and across collections of recordings sharing the same acquisition context rather than subject identity.

\section{Data Curation Pipeline}
\subsection{Data Sources}
\label{sec:datacuration:datasource}
We collect a dataset of about 4M wrist-worn PPG recordings spanning 202K real-world users using smartwatch in daily-life settings.
Depending on acquisition pattern, we group them into two datasets: \textbf{Home Blood Pressure Measurement (HBPM)}, consisting of over 1K users and 1.3M PPG records; and \textbf{Vascular Stiffness (VS)}, consisting of about 200K users and about 3M PPG records.
More dataset details can be found in the supplementary material.

\subsection{PPG Preprocessing and Feature Extraction}
\label{datacuration:data_description}

\noindent\textbf{Band-Pass Filtering:} The raw PPG signal is first preprocessed using a 4th-order Butterworth band-pass filter with cutoff frequencies of 0.25–10 Hz to attenuate baseline drift and high-frequency noise.

\noindent\textbf{Baseline Removal:} To further eliminate slow-varying trends, baseline drift is corrected using an adaptive iteratively reweighted penalized least squares method~\cite{baselineremoval}, yielding a detrended PPG waveform.

\noindent\textbf{Heartbeat Segmentation:} Based on the detrended signal, individual cardiac cycles are segmented using an adaptive extremum-based pulse detection strategy, enabling reliable extraction of beat-level waveforms.

\noindent\textbf{Feature Extraction:}\label{feature_extraction} The extracted beats are subsequently screened to remove low-quality or inconsistent pulses, and hand-crafted features are computed for each remaining beat~\cite{vitalpy}.
The final segment-level representation is obtained by averaging features across all clean beats.

\subsection{Textual Description Construction}
To provide explicit semantic anchors for PPG pretraining, we construct two complementary textual descriptions for each recording \(x_i\): a waveform morphology description \(c_i^{\mathrm{morph}}\) derived from the signal and a metadata-derived context description \(c_i^{\mathrm{ctx}}\) constructed from routinely available attributes.

\subsubsection{Waveform Morphology Description}
To capture morphological characteristics of the waveform, we extract a suite of handcrafted features that characterize the fiducial geometry and temporal dynamics of the PPG waveform.
These features are subsequently serialized into texts via an enumerative template, providing a direct quantitative characterization of the pulse shape and cycle.
An example waveform morphology description $c^{\mathrm{morph}}$ is:
\begin{quote}\small
Systolic upslope proportion 0.16, cardiac cycle 0.96s, 62.76 beats per minute within normal range, 0.32s duration systolic-diastolic peak interval, relaxation 0.58s, ventricular contraction 0.38s.
\end{quote}

\subsubsection{Metadata-Derived Context Description}
\label{sec:datacuration:physiology_caption}
For each PPG recording \(x_i\), the available metadata includes age, sex, height, weight and acquisition time.
We additionally compute BMI and its categorical body-type descriptor from height and weight.
The discretized context key is defined as $a_i = (a_i^{\mathrm{age}}, a_i^{\mathrm{sex}}, a_i^{\mathrm{BMI}}, a_i^{\mathrm{time}})$, which is used to aggregate recordings with identical context keys during Cross-Recording Context Alignment.
Then we design a template-based mapping function to convert the metadata context into a a semantically coherent natural-language description \(c_i^{\mathrm{ctx}}\).
Specifically, numerical values (age, weight, etc.) are projected into a predefined template, where raw quantitative values are complemented by corresponding qualitative descriptors.
Detailed projection rules are provided in the supplementary material.
An example metadata-derived context description $c^{\mathrm{ctx}}$ is:
\begin{quote}\small
A 36-year-old young woman with a normal weight figure, with a height of 158 cm, 52 kg in weight, with data collected at morning 8:19.
\end{quote}



\section{Method}
\label{sec:method}

\subsection{Overview}

Given the curated PPG-Text corpus, $x_i$ is paired with a morphology description $c_i^{\mathrm{morph}}$, a context description $c_i^{\mathrm{ctx}}$, and a metadata tuple $a_i$.
A PPG encoder and a pretrained text encoder map the waveforms and descriptions into a shared embedding space. 
For simplicity, we use $h_i^p$, $h_i^{\mathrm{morph}}$, and $h_i^{\mathrm{ctx}}$ to denote the resulting PPG and text embeddings.

As illustrated in Fig.\ref{fig:method:overview}, \method jointly optimizes three complementary objectives.
Record-Level Morphology Alignment treats each PPG recording as an individual alignment unit and grounds it in its waveform description.
Cross-Recording Context Alignment changes the PPG-side alignment unit from one recording to a set of recordings sharing the same metadata tuple.
Waveform-Guided Text Modeling complements these global alignment objectives by conditioning the recovery of masked contextual concepts on PPG patch representations.

\subsection{Record-Level Morphology Alignment}

This objective aligns each PPG recording with the morphology description constructed from its hand-crafted waveform features.
For a batch of $B$ recordings, $h_i^p$ is aligned with its paired description $h_i^{\mathrm{morph}}$:
\begin{align*}
    \mathcal{L}_{\mathrm{morph}}
    =&-\frac{1}{2B}\sum_{i=1}^{B}\log
    \frac{\exp\!\left(s(h_i^p,h_i^{\mathrm{morph}})/\tau\right)}
    {\sum_{j=1}^{B}\exp\!\left(s(h_i^p,h_j^{\mathrm{morph}})/\tau\right)}
    \\
    &-\frac{1}{2B}\sum_{i=1}^{B}\log
    \frac{\exp\!\left(s(h_i^p,h_i^{\mathrm{morph}})/\tau\right)}
    {\sum_{j=1}^{B}\exp\!\left(s(h_j^p,h_i^{\mathrm{morph}})/\tau\right)}
\end{align*}

where $s(\cdot,\cdot)$ denotes cosine similarity and $\tau$ is the learnable temperature.
Since one recording constitutes one alignment unit, this objective preserves record-specific morphological information in the PPG representation.

\subsection{Cross-Recording Context Alignment}
Within a training batch $\mathcal{B}$, samples are partitioned into $M$ groups sharing the same acquisition context (i.e. the same metadata tuple): $\mathcal{B}=\bigcup_a\mathcal{G}_a$, where $\mathcal{G}_a=\{x\in\mathcal{B}\mid a(x)=a\}$.
The context alignment objective is designed to anchor the shared part of PPG representations to their corresponding context descriptions.

For each group $\mathcal{G}_a$, we aggregate the corresponding recording embeddings via learnable attention query:
$$h_{a_m}^p = \text{AttentionAggregate}(\{h_i^p\mid i\in\mathcal{G}_a\})$$
and the context description embeddings are randomly sampled from the group.
Then the context-level alignment objective is defined as:
\begin{align*}
    \mathcal{L}_{\mathrm{ctx}}
    =&-\frac{1}{2M}\sum_{m=1}^{M}\log
    \frac{\exp\!\left(s( h_{a_m}^p, h_{a_m}^{\mathrm{ctx}})/\tau\right)}
{\sum_{n=1}^{M}
 \exp\!\left(s( h_{a_m}^p, h_{a_n}^{\mathrm{ctx}})/\tau\right)}
\\
    &-\frac{1}{2M}\sum_{m=1}^{M}\log
    \frac{\exp\!\left(s( h_{a_m}^p, h_{a_m}^{\mathrm{ctx}})/\tau\right)}
{\sum_{n=1}^{M}
 \exp\!\left(s( h_{a_n}^p, h_{a_m}^{\mathrm{ctx}})/\tau\right)}
\end{align*}

By aligning a context-defined recording group with its associated description, the anchored PPG representations capture the contextual regularities observed across comparable recordings.

\subsection{Waveform-Guided Text Modeling}
\begin{table*}[!t]
    \centering
    \small
    \setlength{\tabcolsep}{4pt}
    \begin{tabular}{l|cc|cc|cc}
        \toprule
        \multirow{2}{*}{Method}
        & \multicolumn{2}{c|}{SBP}
        & \multicolumn{2}{c|}{DBP}
        & \multicolumn{2}{c}{cfPWV} \\
        & MAE\(\downarrow\) & \(R^2\uparrow\)
        & MAE\(\downarrow\) & \(R^2\uparrow\)
        & Balanced ACC\(\uparrow\) & PR-AUC\(\uparrow\) \\ \midrule
        SimCLR~\cite{simclr}            & 11.89\(\pm\)0.69 & 0.09\(\pm\)0.04  & 6.71\(\pm\)0.13 & 0.09\(\pm\)0.04 & 0.84\(\pm\)0.03 & 0.80\(\pm\)0.04 \\
        MoCo-v3~\cite{mocov3}           & 11.98\(\pm\)0.73 & 0.07\(\pm\)0.04  & 6.67\(\pm\)0.23 & 0.08\(\pm\)0.04  & \underline{0.85\(\pm\)0.05} & 0.80\(\pm\)0.06 \\
        MAE~\cite{mae}                  & 11.95\(\pm\)0.47 & 0.07\(\pm\)0.05  & 6.81\(\pm\)0.14 & 0.05\(\pm\)0.06  & 0.84\(\pm\)0.03 & 0.80\(\pm\)0.03 \\
        PaPaGei~\cite{papagei}          & 12.42\(\pm\)0.48 & 0.00\(\pm\)0.05  & 6.94\(\pm\)0.31 & 0.00\(\pm\)0.04  & 0.83\(\pm\)0.05 & 0.79\(\pm\)0.05 \\
        Apple~\cite{apple0}             & 12.01\(\pm\)0.84 & 0.07\(\pm\)0.06  & 6.67\(\pm\)0.28 & 0.09\(\pm\)0.05  & 0.85\(\pm\)0.03 & 0.79\(\pm\)0.03 \\ \midrule
        SupCon~\cite{khosla2020supcon}  & 11.94\(\pm\)0.51 & 0.06\(\pm\)0.04  & 6.78\(\pm\)0.13 & 0.07\(\pm\)0.04  & 0.77\(\pm\)0.07 & 0.80\(\pm\)0.05 \\
        Metadata Embedding              & 12.18\(\pm\)0.50 & 0.05\(\pm\)0.03  & 6.88\(\pm\)0.19 & 0.03\(\pm\)0.03  & 0.78\(\pm\)0.04 & 0.75\(\pm\)0.05 \\
        Multi-Task Supervision          & 12.67\(\pm\)0.51 & -0.03\(\pm\)0.04 & 7.25\(\pm\)0.17 & -0.06\(\pm\)0.04 & 0.83\(\pm\)0.02 & 0.81\(\pm\)0.03 \\ \midrule
        Morphology Alignment               & 11.92\(\pm\)0.50 & 0.08\(\pm\)0.04  & 6.66\(\pm\)0.12 & 0.07\(\pm\)0.03  & 0.84\(\pm\)0.02 & 0.76\(\pm\)0.03 \\
        + Context Alignment             & \underline{11.73\(\pm\)0.48} & \underline{0.10\(\pm\)0.04} & \underline{6.60\(\pm\)0.20} & \underline{0.11\(\pm\)0.04} & 0.83\(\pm\)0.03 & \underline{0.83\(\pm\)0.04} \\
        + WGTM (\method)                & \textbf{11.67\(\pm\)0.48} & \textbf{0.11\(\pm\)0.03} & \textbf{6.57\(\pm\)0.11} & \textbf{0.11\(\pm\)0.05} & \textbf{0.85\(\pm\)0.02} & \textbf{0.84\(\pm\)0.03} \\ \bottomrule
    \end{tabular}
    \caption{Linear-probing transfer to systolic blood pressure (SBP) and diastolic blood pressure (DBP) regression on HBPM-X, and high-stiffness arterial classification on cfPWV. Results are mean\(\pm\)std under 5 random seeds. Best results are in \textbf{bold}, and second-best results are \underline{underlined}.}
    \label{tab:main_probe_bp_pwv}
\end{table*}

The two alignment objectives operate on global representations and do not explicitly require individual contextual concepts to be supported by waveform tokens.
We therefore mask selected attributes in the context description and condition their prediction on the PPG patch representations $H_i^p$ through cross-attention.

Let $\mathcal{M}_i$ denote the masked token positions and
$c_{i,\setminus\mathcal{M}_i}^{\mathrm{ctx}}$ the remaining context description.
The Waveform-Guided Text Modeling loss is
\begin{align*}
    \mathcal{L}_{\mathrm{wgtm}}
    = -\sum_{i} \sum_{k\in\mathcal{M}_i} \log P\left( w_{i,k} \mid c_{i,\setminus\mathcal{M}_i}^{\mathrm{ctx}}, H_i^p \right)
    \label{eq:wgtm}
\end{align*}
where $w_{i,k}$ is the original token at position $k$.
This objective provides a token-level complement to global PPG-Text
alignment by encouraging contextual concepts to be recoverable from
the waveform.

\method jointly optimizes the three complementary losses:
\begin{align*}
    \mathcal{L}
    =\lambda_{\mathrm{morph}}\mathcal{L}_{\mathrm{morph}}
    +\lambda_{\mathrm{ctx}}\mathcal{L}_{\mathrm{ctx}}
    +\lambda_{\mathrm{wgtm}}\mathcal{L}_{\mathrm{wgtm}}.
\end{align*}

\section{Experiments}
\subsection{Research Questions and Baselines.}
We organize our experiments around three research questions.
\textbf{RQ1: Transferability.}
Does \method learn transferable PPG representations across physiological tasks and datasets?
\textbf{RQ2: Multi-scope textual supervision.}
Does matching each supervision source to its reliable waveform-evidence scope and expressing metadata through language improve representation learning over alternative uses of the same metadata?
\textbf{RQ3: Semantic accessibility.}
Does multi-scope text alignment make contextual physiological semantics directly accessible from PPG without task-specific training?

For RQ1, we compare \method with representative contrastive, reconstruction-based, and PPG-specific pretraining methods, including SimCLR~\cite{simclr}, MoCo-v3~\cite{mocov3}, MAE~\cite{mae}, Apple~\cite{apple0}, and PaPaGei~\cite{papagei}.
For RQ2, we design three targeted baselines using the same contextual metadata as \method.
SupCon~\cite{khosla2020supcon} treats recordings sharing the same metadata tuple as positives under record-level supervised contrastive learning.
Multi-Task Supervision directly predicts age, sex, BMI, and acquisition time through separate task heads.
Metadata Embedding retains metadata-grouped PPG aggregation and contrastive alignment, but replaces contextual language with an embedding learned directly from one-hot encoded attribute vectors.
We additionally compare three cumulative ablation variants.
For RQ3, we compare the three cumulative variants in zero-shot user profiling.
Random prediction provides a chance-level reference, while a task-specific supervised model provides a downstream-trained reference.
\paragraph{Implementation Details.}
\method is pretrained on HBPM-BG and VS.
We employ transformer as PPG encoder with feature dimension 512.
For text encoder, we initialize it from PubMedBert~\cite{pubmedbert}, which is pretrained on massive PubMed corpus.
PubMedBert is composed of 12 layers of transformer blocks and we choose the bottom 6 layers as text encoder and  insert a cross-attention module in each layer of the top 6 transformer layers to form the text decoder~\cite{albef}.
We set the batchsize as 512 and learning rate as 1e-4 with 1000 steps of linear warmup and decays in cosine schedule for pretraining \method for 100 epochs on 4 NVIDIA A100-80G.
Unless otherwise stated, the pretrained PPG encoder is frozen and only a task-specific linear predictor is optimized.
We set three loss weights to 1.0.

\subsection{Linear-Probe Transfer}
\begin{table*}[!t]
    \centering
    \small
    \setlength{\tabcolsep}{4pt}
    \begin{tabular}{l|cc|cc|cc}
        \toprule
        \multirow{2}{*}{Method}
        & \multicolumn{2}{c|}{Age}
        & \multicolumn{2}{c|}{Sex}
        & \multicolumn{2}{c}{HR} \\
        & MAE\(\downarrow\) & \(R^2\uparrow\)
        & ACC\(\uparrow\) & ROC-AUC\(\uparrow\)
        & MAE\(\downarrow\) & \(R^2\uparrow\) \\ \midrule
        SimCLR~\cite{simclr}            & 7.68\(\pm\)0.89 & 0.34\(\pm\)0.11 & 0.68\(\pm\)0.07 & 0.78\(\pm\)0.08 & 8.87\(\pm\)3.46 & 0.44\(\pm\)0.30 \\
        MoCo-v3~\cite{mocov3}           & 7.73\(\pm\)0.83 & 0.34\(\pm\)0.11 & 0.66\(\pm\)0.09 & 0.77\(\pm\)0.09 & 9.24\(\pm\)3.83 & 0.40\(\pm\)0.34 \\
        MAE~\cite{mae}                  & \underline{7.08\(\pm\)0.82} & 0.41\(\pm\)0.11 & \underline{0.74\(\pm\)0.05} & 0.80\(\pm\)0.08 & 8.36\(\pm\)3.11 & 0.44\(\pm\)0.37 \\
        PaPaGei~\cite{papagei}          & 8.83\(\pm\)0.74 & 0.19\(\pm\)0.11 & 0.66\(\pm\)0.07 & 0.73\(\pm\)0.08 & 8.29\(\pm\)3.53 & 0.43\(\pm\)0.42 \\
        Apple~\cite{apple0}             & 7.53\(\pm\)0.82 & 0.37\(\pm\)0.10 & 0.68\(\pm\)0.07 & 0.79\(\pm\)0.07 & 8.18\(\pm\)3.44 & \textbf{0.50\(\pm\)0.39} \\ \midrule
        SupCon~\cite{khosla2020supcon}  & 7.44\(\pm\)0.62 & 0.37\(\pm\)0.09 & \underline{0.72\(\pm\)0.06} & \underline{0.81\(\pm\)0.09} & 8.22\(\pm\)3.09 & 0.49\(\pm\)0.31 \\
        Metadata Embedding              & 7.62\(\pm\)0.71 & 0.37\(\pm\)0.08 & 0.70\(\pm\)0.06 & 0.80\(\pm\)0.09 & 13.09\(\pm\)4.57 & -0.00\(\pm\)0.55 \\
        Multi-Task Supervision          & 7.32\(\pm\)0.87 & 0.38\(\pm\)0.12 & 0.70\(\pm\)0.07 & 0.80\(\pm\)0.07 & 9.49\(\pm\)3.10 & 0.36\(\pm\)0.38 \\ \midrule
        Morphology Alignment               & 7.56\(\pm\)0.78 & 0.37\(\pm\)0.10 & 0.70\(\pm\)0.04 & 0.80\(\pm\)0.07 & 8.11\(\pm\)3.46 & 0.44\(\pm\)0.38 \\
        + Context Alignment
        & 7.10\(\pm\)0.85 & \underline{0.42\(\pm\)0.11} & 0.74\(\pm\)0.07 & 0.82\(\pm\)0.08 & \underline{7.92\(\pm\)3.48} & 0.46\(\pm\)0.41 \\
        + WGTM (\method)               & \textbf{7.05\(\pm\)0.79} & \textbf{0.42\(\pm\)0.11} & \textbf{0.74\(\pm\)0.06} & \textbf{0.83\(\pm\)0.08} & \textbf{7.65\(\pm\)3.09} & \underline{0.49\(\pm\)0.36} \\ \bottomrule
    \end{tabular}
    \caption{Linear-probing results for attribute probing on HBPM-X and external heart-rate transfer on PPG-DaLiA dataset. Best results are in \textbf{bold}, and second-best results are \underline{underlined}.}
    \label{tab:main_probe_age_sex_hr}
\end{table*}
\begin{table}[!t]
    \centering
    \small
    \setlength{\tabcolsep}{1mm}
    \begin{tabular}{lcccc}
        \toprule
        Method & Age Group & Sex & BMI & Joint Acc. \\ \midrule
        Random & 0.2591 & 0.4860 & 0.2384 & 0.0384 \\
        Supervised & \textbf{0.7979} & 0.6610 & 0.4056 & 0.2136 \\ \midrule
        Morphology Alignment
        & 0.2749 & 0.4881 & 0.3514 & 0.0379 \\
        + Context Alignment
        & 0.6272 & \underline{0.8189} & \underline{0.5133} & \underline{0.2560} \\
        + WGTM (\method)
        & \underline{0.7078} & \textbf{0.8313} & \textbf{0.5350} & \textbf{0.2955} \\ \bottomrule
    \end{tabular}
    \caption{Zero-shot user-profiling accuracy for age group, sex, BMI, and their joint prediction. The supervised row is a task-specific reference trained with downstream labels. Best results are in \textbf{bold}, and second-best results are \underline{underlined}.}
    \label{tab:experiment:zeroshotradar}
\end{table}

\subsubsection{Evaluation datasets.}
We evaluate the pretrained PPG encoder on six downstream tasks from three datasets.
HBPM-X provides recording-level SBP and DBP regression tasks and subject-level age regression and sex classification.
The cfPWV cohort is used for subject-level arterial-stiffness classification, where predictions from multiple recordings of the same subject are averaged before evaluation.
Cross-dataset heart-rate transfer is evaluated on PPG-DaLiA~\cite{ppg3} using the standard 15-subject leave-one-subject-out protocol, differing from other tasks.
We use MAE as the primary metric for regression, PR-AUC for imbalanced cfPWV classification, and ROC-AUC for sex classification.

\subsubsection{RQ1: Transferability.}
For clinical cardiovascular transfer in Table~\ref{tab:main_probe_bp_pwv}, \method achieves the strongest MAEs of 11.67 and $6.57$ for SBP and DBP estimation.
It also achieves the best performance on Balanced Accuracy of 0.85 and PR-AUC of 0.84 for imbalanced cfPWV classification, exceeding the best comparison result of 0.83.
For subject-level age regression and sex classification in Table~\ref{tab:main_probe_age_sex_hr}, \method achieves the strongest MAE of 7.05 and AUC of 0.83 for age and sex, respectively.
On the external PPG-DaLiA dataset, \method reduces heart-rate MAE from the previous best of 8.18 to 7.65, while obtaining a comparable $R^2$ of 0.49.
The consistent performance across blood pressure, vascular stiffness, physiological attributes, and an external heart-rate dataset indicates that the learned representation is not specialized to a single type of downstream target.

\subsubsection{RQ2: Multi-scope textual supervision.}
Comparing with conventional waveform-only methods, Morphology Alignment already provides a competitive starting point, which is consistent with its intended role in waveform morphology grounding.
Starting from this morphology-grounded representation, adding Context Alignment improves all primary downstream metrics and constitutes the largest structural gain among the cumulative variants.
For example, age MAE decreases from 7.56 to 7.10, while cfPWV PR-AUC increases from 0.76 to 0.83.
The context-aligned variant also outperforms SupCon on all primary metrics.
Both methods use the same metadata-defined relations but context-level alignment constructs a more reliable supervision by integrating a broader scope of PPG recordings, which demonstrates the effectiveness of the proposed multi-scope supervision.

Comparing with Multi-Task Supervision, the consistent performance gain of context-aligned variant suggests that directly predicting metadata encourages the encoder to retain information sufficient for the predefined outputs, but does not necessarily produce a broadly transferable representation space.
In contrast, Context Alignment uses the joint metadata tuple to define the context under which physiological regularities are more observable.
Metadata therefore acts as an organizing principle of supervision rather than as a set of isolated prediction endpoints.
This allows the encoder to preserve a broader range of context-associated waveform variation that remains useful beyond the metadata tasks themselves.

Metadata Embedding provides a further control on the usage of metadata.
It retains the same grouping and aggregating strategy as Context Alignment, but replaces contextual descriptions with one-hot attribute vectors processed by a MLP.
In comparison, a one-hot representation initially treats metadata values as unrelated symbols and requires the embedding to learn their geometry solely from the alignment objective.
The gain from contextual description is therefore not only explained by metadata-based grouping but also by a non-arbitrary linguistic and semantic structure offered by the pretrained text encoder.

Finally, WGTM functions on a complementary design dimension.
Cross-Recording Context Alignment changes the scope of waveform evidence from an individual recording to a metadata-defined collection, whereas WGTM changes the resolution of supervision from global representation alignment to patch-level correspondence.
Adding WGTM to Context Alignment produces smaller but consistent improvement.
They indicate that fine-grained waveform-conditioned concept recovery enriches the representation beyond coarse global alignment, further improving its physiological content and downstream transferability.

\subsection{Zero-shot User Profiling}
\label{sec:exp:zeroshot}
Beyond conventional transfer learning, \method enables contextual information to be queried directly by language through the aligned representation space.
We introduce \textbf{Zero-shot User Profiling}, a new task aiming at inferring age group, sex, and BMI category of a subject from its single or multiple PPG recordings without optimizing a downstream predictor.

For each subject, multiple PPG recordings are encoded and the resulting representations are aggregated using the learned attention query vector.
To construct the candidate text representations, we enumerate all valid combinations of age group, sex, BMI category, and acquisition time and instantiate them using the contextual description template.
We retrieve the top-$k$ candidates and predict age group, sex, and BMI category by field-wise voting.
The acquisition time participates in candidate construction but is marginalized during prediction.

\subsubsection{RQ3: Semantic accessibility.}
In Table~\ref{tab:experiment:zeroshotradar}, we use 5 random seeds to report the mean profiling accuracy with $k=3$, beyond which performance shows limited variation.
Morphology Alignment variant performs slightly better than the random reference, which is expected.
Adding Cross-Recording Context Alignment yields substantial improvements across all evaluated fields, demonstrating that the metadata-defined alignment objective makes context-related semantics directly accessible through language.
The complete \method performance proves the critical role of fine-grained patch-level supervision.

\begin{figure}[th]
    \centering
    \includegraphics[width=0.6\linewidth]{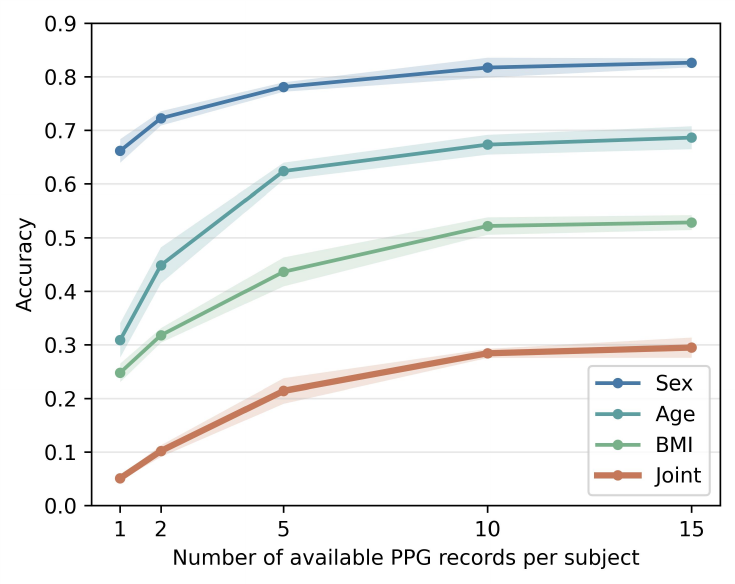}
    \caption{Zero-shot user-profiling accuracy for different number of PPG recordings used to construct aggregated representation for a single subject.}
    \label{fig:zero_shot_ppg_num}
\end{figure}

To further examine the role of cross-recording aggregation, we vary the number of PPG recordings used to construct aggregated representation for a single subject among $1, 2, 5, 10, 15$.
Fig.~\ref{fig:zero_shot_ppg_num} shows that profiling accuracy increases monotonically as more recordings are aggregated.
Because the improvement from 10 to 15 recordings is limited, we report the 10-record setting in Table~\ref{tab:experiment:zeroshotradar}.
This result further validates that physiological regularities are weakly and indirectly reflected by a single PPG recording and emphasizes the necessity of constructing supervision on a broader scope rather than individually during PPG representation learning.

\begin{figure}[thbp]
  \centering
  \begin{subfigure}[b]{0.48\linewidth}
    \centering
    \includegraphics[width=\linewidth]{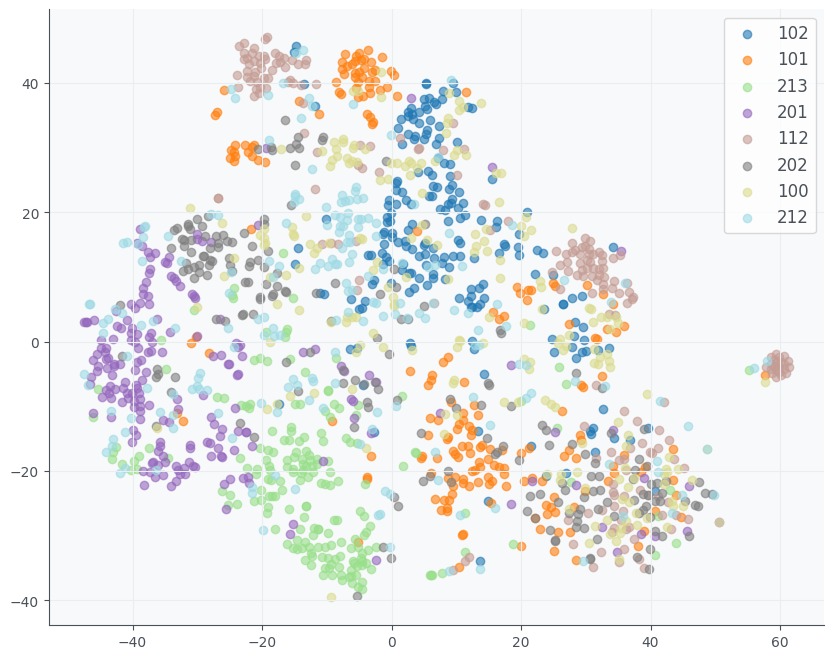}
    \caption{Apple}
    \label{fig:exp:tsne_baseline}
  \end{subfigure}
  \hfill
  \begin{subfigure}[b]{0.48\linewidth}
    \centering
    \includegraphics[width=\linewidth]{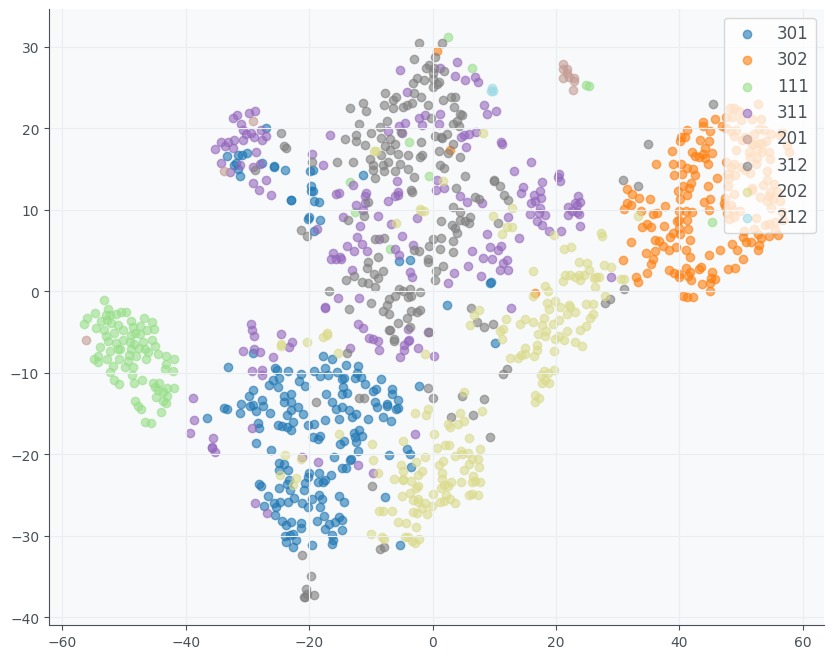}
    \caption{\method(Ours)}
    \label{fig:exp:tsne_ours}
  \end{subfigure}
  
  \caption{T-SNE visualization of two methods on the HBPM-X. Users with the same attribute tuple \(a_s\) are marked in the same color.}
  \label{fig:exp:tsne}
\end{figure}

\subsection{Additional Qualitative Results}
\subsubsection{UniModal Retrieval Analysis}
\label{sec:exp:unimodalretrieval}
\begin{figure}[th]
    \centering
    \includegraphics[width=\linewidth]{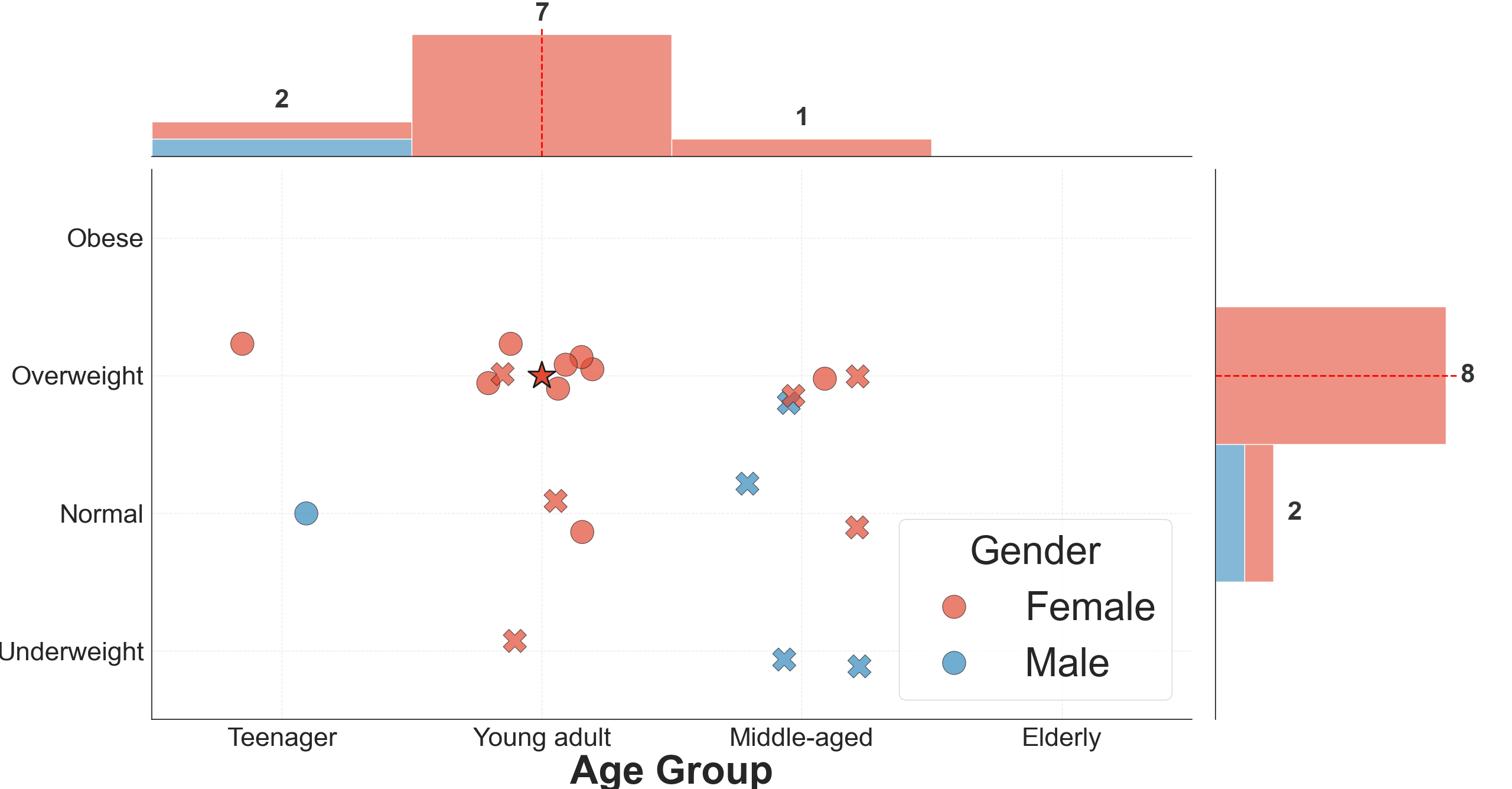}
    \caption{Visualization of unimodal retrieval. The red star denotes the ground truth attribute tuple of the query PPG, and we use circle to plot the attribute distribution of recalled PPG signals(Age group distribution on the top and BMI group distribution on the right). We also visualize the result by Apple in the X-label.}
    \label{fig:exp:retrieval_analysis}
\end{figure}

To verify whether the single PPG encoder \(f_{\theta}\) captures physiological semantics, we conduct a unimodal signal-to-signal retrieval experiment.
Specifically, for a given query PPG signal, we retrieve the top-10 nearest neighbors from the dataset (excluding the query subject itself) and compare physiological attributes of these signals with the query's ground truth.
As illustrated in Fig.\ref{fig:exp:retrieval_analysis}, for a query subject identified as (Young adult, woman, overweight), the recalled signals(circle) exhibit remarkable semantic clustering, achieving accuracy of $70\%$ in age, $80\%$ in BMI, and $90\%$ in sex.
In contrast, the signals recalled by baseline method Apple~\cite{apple0} do not exhibit the attribute consistency. This implies that \method PPG encoder \(f_{\theta}\) successfully maps physiologically similar signals to adjacent regions under the guidance of text encoder \(g_{\phi}\).

\subsubsection{T-SNE Visualization}

We visualize the representations on HBPM-X using t-SNE and compare \method with baseline method Apple. As illustrated in Fig.\ref{fig:exp:tsne_ours}, while the baseline tends to cluster signals based on individual identity and fails to distinguish between physiology tuples, \method successfully groups them based on shared physiological semantics. This confirms that \method prioritizes not only low-level morphological variations but also high-level physiological semantics.

\section{Conclusion}
%
%


Wearable PPG contains different types of physiological regularities that require different scopes of waveform evidence to be reliably represented. While local morphology is directly observable within an individual recording, context-related regularities can be weakly expressed or obscured by transient factors. \method implements this evidence-scope matching principle by constructing and aligning supervision at complementary scopes. Its consistent performance across downstream transfer and zero-shot profiling shows that this design produces reusable and semantically accessible PPG representations. This work suggests that considering where supervisory evidence becomes reliable provides a useful perspective for representation learning of wearable biosignals.

\clearpage
\section*{Appendix}
\appendix
\section{Dataset Details}
\label{sec:appendix:data_details}

\subsection{Pretraining Datasets}

Because this work studies wearable PPG representation learning, pretraining requires wrist-worn PPG collected in daily-life settings rather than short clinical recordings alone.
It further needs dense repeated observations per user, together with age, sex, height, weight, and acquisition timestamps, so that recordings can be grouped by shared metadata tuples.
HBPM also provides longitudinally repeated cuff-synchronized PPG--BP pairs, and the corpus must be large enough to cover many context combinations.
Existing public PPG datasets typically lack this combination of wearable daily-life coverage, within-user sampling density, complete contextual metadata, user scale, and clinical labels.
We therefore use the internal HBPM and VS corpora described below.

\paragraph{HBPM-BG.}
The Home Blood Pressure Measurement (HBPM) study lasted 30 days.
In addition to scheduled paired PPG and BP measurements, the smartwatch automatically acquired one-minute background PPG recordings approximately every 30 minutes.
These unlabeled recordings form the HBPM-BG pretraining set, which contains approximately 1.3 million PPG recordings from 1{,}086 users.
HBPM-BG has no user overlap with the downstream HBPM-X cohort.

\paragraph{Vascular Stiffness (VS).}
The VS pretraining set was collected from a deployed smartwatch function with user consent.
Recordings were acquired automatically when the inertial sensors indicated that the wearer was stationary.
VS contains approximately 3 million one-minute PPG recordings from 200.5K users and has no user overlap with either HBPM-BG or HBPM-X.

\begin{figure*}[t]
    \centering
    \includegraphics[width=0.75\linewidth]{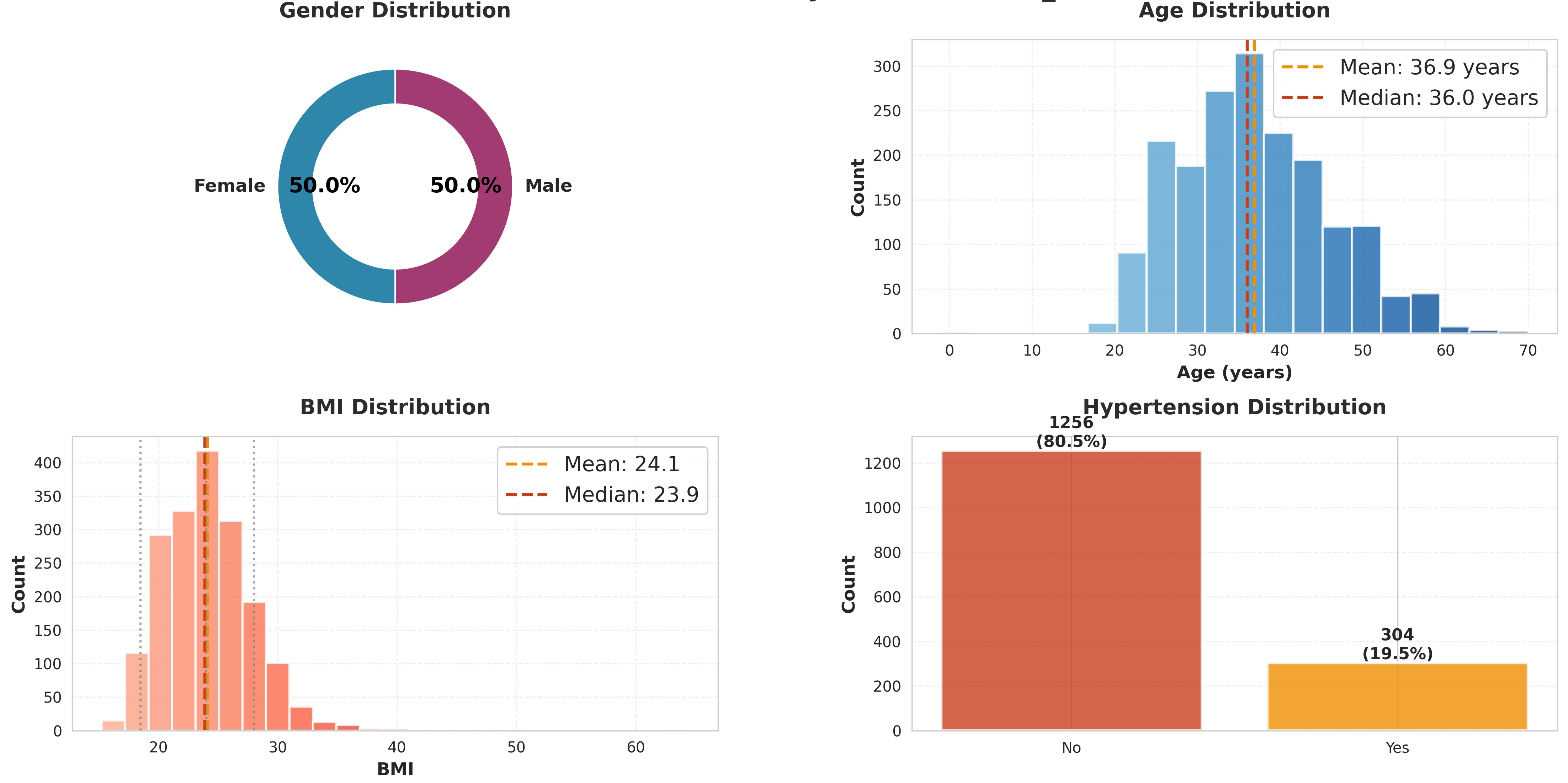}
    \caption{User-level attribute distribution of the HBPM dataset.}
    \label{fig:appendix:hbpm_users}
\end{figure*}

\begin{figure*}[t]
    \centering
    \includegraphics[width=0.75\linewidth]{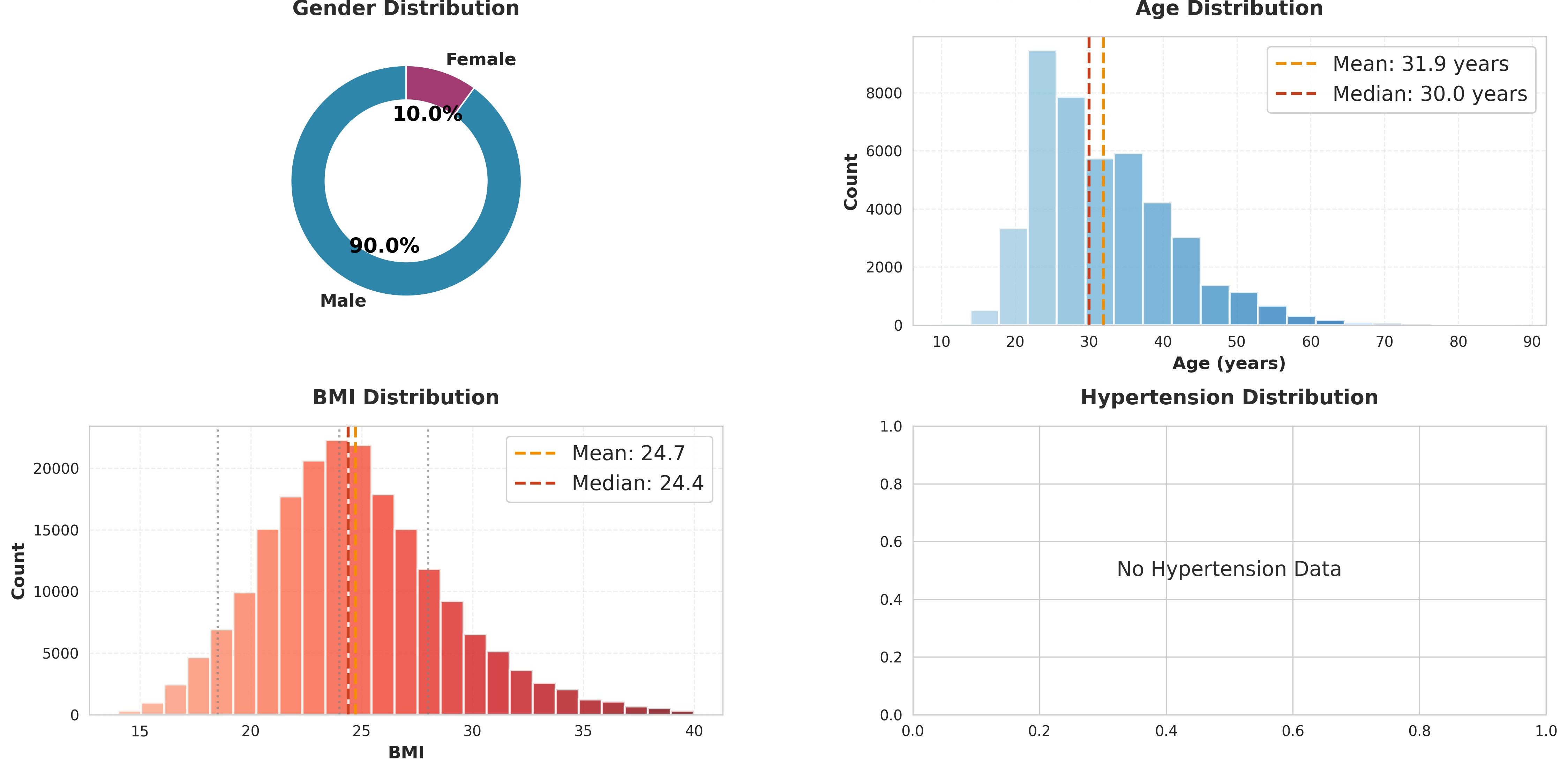}
    \caption{User-level attribute distribution of the VS dataset.}
    \label{fig:appendix:vs_users}
\end{figure*}

All HBPM-BG and VS recordings were acquired at 250~Hz in daily-life settings.
Their user-level attribute distributions are shown in Fig.~\ref{fig:appendix:hbpm_users} and Fig.~\ref{fig:appendix:vs_users}.

\subsection{Downstream Datasets}

\paragraph{HBPM-X.}
For scheduled measurements in the HBPM study, participants rested, recorded one minute of wrist PPG, and measured reference blood pressure using an OMRON U36T digital cuff monitor.
The protocol was repeated during morning, afternoon, and evening sessions, producing at least nine paired measurements per participant per day.
HBPM-X contains approximately 90{,}000 PPG and BP pairs from 303 users.
We use it for systolic blood pressure (SBP) regression, diastolic blood pressure (DBP) regression, age regression, sex classification, and zero-shot user profiling.

\begin{figure*}[t]
    \centering
    \includegraphics[width=\linewidth]{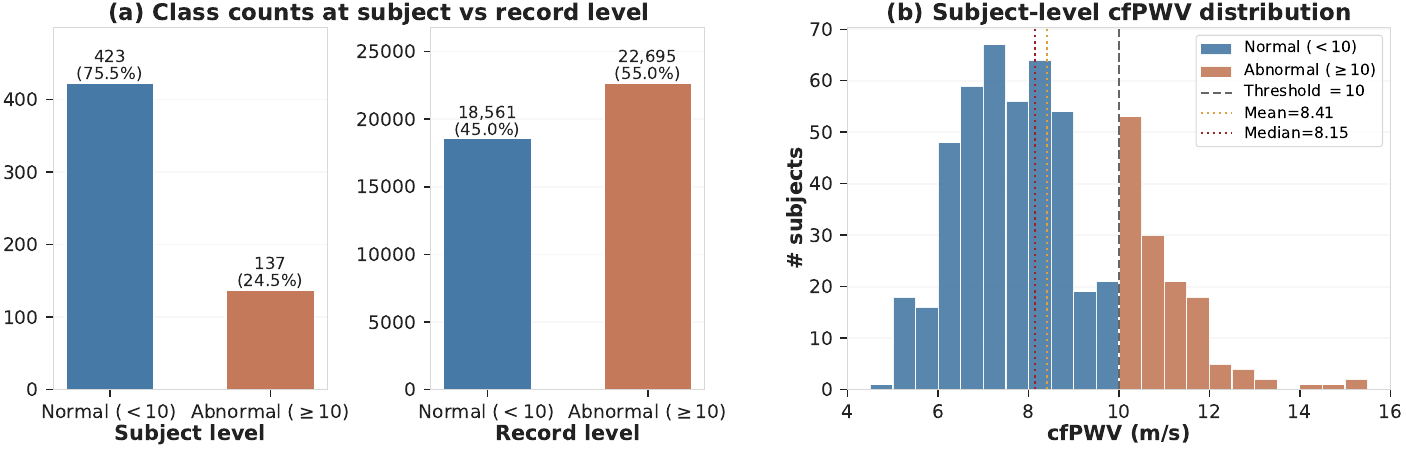}
    \caption{cfPWV cohort label statistics. (a) Class counts at the subject and record levels, revealing inverted imbalance. (b) Subject-level continuous cfPWV distribution with the clinical threshold at $10\,\mathrm{m/s}$.}
    \label{fig:appendix:cfpwv_ab}
\end{figure*}

\begin{figure*}[t]
    \centering
    \includegraphics[width=0.85\linewidth]{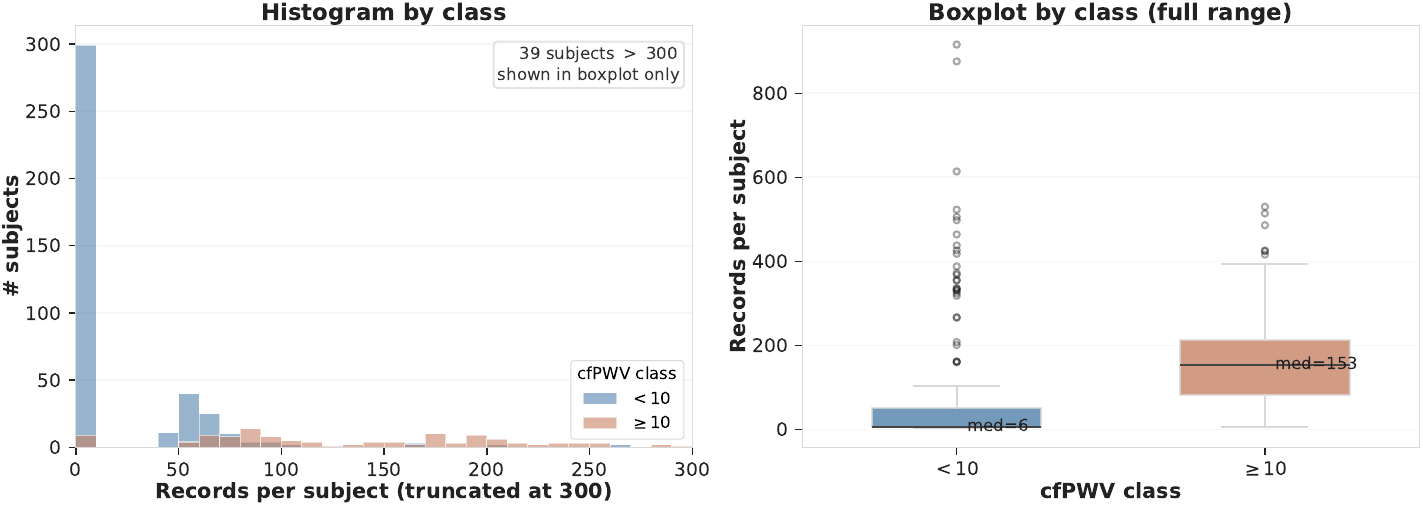}
    \caption{Records-per-subject distribution in the cfPWV cohort. The histogram is truncated at 300 for readability; the boxplot shows the full range. Abnormal subjects contribute substantially more recordings on average.}
    \label{fig:appendix:cfpwv_c}
\end{figure*}

\paragraph{cfPWV.}
The cfPWV cohort contains 41{,}256 single-channel PPG recordings from 560 subjects.
Elevated arterial stiffness is defined by carotid--femoral pulse wave velocity $\mathrm{cfPWV} \geq 10.0\,\mathrm{m/s}$~\cite{vanbortel2012cfpwv}.
At the subject level, 423 subjects are in the normal class and 137 are in the abnormal class.
We use this cohort for subject-level binary classification.
Fig.~\ref{fig:appendix:cfpwv_ab} summarizes the class counts and continuous cfPWV distribution, and Fig.~\ref{fig:appendix:cfpwv_c} shows that the number of records per subject is long-tailed and class-dependent.
Because of this structure, we evaluate at the subject level after averaging record-level logits within each subject.

\paragraph{PPG-DaLiA.}
PPG-DaLiA~\cite{ppg3} is a public wrist-PPG dataset containing recordings from 15 subjects.
The official heart-rate targets are defined over 8-second windows with a 2-second stride.
For each target window, we extract a 12-second PPG segment centered on that window and resample the signal from 60 Hz to 250 Hz.
Only windows that can be fully extracted from the continuous recording are retained.

\section{Description Generation}
\subsection{Morphology Features}
\label{sec:appendix:features}
Table~\ref{tab:appendix:features} lists the beat-level features used to construct waveform morphology descriptions.
Features are averaged over clean beats within each recording and rounded to two decimal places before text serialization.

\begin{table*}[t]
    \centering
    \begin{tabular*}{0.6\textwidth}{@{\extracolsep{\fill}}cc@{}}
        \toprule
        Symbol & Definition \\
        \midrule
        $T$ & Pulse-cycle duration (s) \\
        $\mathrm{IHR}$ & Instantaneous heart rate (bpm) \\
        $\mathrm{prop\_s}$ & Fraction of the cycle spent on the systolic upslope \\
        $\Delta T$ & Interval between systolic and diastolic peaks (s) \\
        $t_{\mathrm{sys}}$ & Systolic duration (s) \\
        $t_{\mathrm{dia}}$ & Diastolic duration (s) \\
        $\mathrm{AW}_s$ & Amplitude-to-width ratio of the systolic peak \\
        $\mathrm{AW}_d$ & Amplitude-to-width ratio of the diastolic peak \\
        \bottomrule
    \end{tabular*}
    \caption{Handcrafted PPG features used to curate morphology descriptions.}
    \label{tab:appendix:features}
\end{table*}

\subsection{Caption Templates}
\label{sec:appendix:caption_templates}

Each pretraining sample is paired with a morphology description $c^{\mathrm{morph}}$ and a context description $c^{\mathrm{ctx}}$.
Both are rendered from deterministic slot templates rather than free-form generation.

\paragraph{Morphology description.}
We serialize the eight features in Table~\ref{tab:appendix:features} into five clause groups---Cycle, SysDia, Prop, DeltaT, and AW---and form each morphology caption $c^{\mathrm{morph}}$ by concatenating a fixed prefix with a uniformly shuffled ordering of these groups.
All five groups are retained for each recording.
Within a group, surface wording is sampled from a small paraphrase set and clause order is shuffled uniformly across groups.
Table~\ref{tab:appendix:morph_slots} summarizes the clause inventory.

\begin{table*}[t]
    \centering
    \begin{tabular*}{0.7\textwidth}{@{\extracolsep{\fill}}ccc@{}}
        \toprule
        Clause & Slots & Example surface form \\
        \midrule
        Cycle & $T$, IHR & cardiac cycle 0.96s, 62.76 bpm within normal range \\
        Sys/Dia & $t_{\mathrm{sys}}$, $t_{\mathrm{dia}}$ & ventricular contraction 0.38s, relaxation 0.58s \\
        Prop & $\mathrm{prop\_s}$ & systolic upslope proportion 0.16 \\
        DeltaT & $\Delta T$ & 0.32s duration systolic-diastolic peak interval \\
        AW & $\mathrm{AW}_s$, $\mathrm{AW}_d$ & amplitude to width ratio: systolic 1.20, diastolic 0.80 \\
        \bottomrule
    \end{tabular*}
    \caption{Clause inventory for morphology descriptions.}
    \label{tab:appendix:morph_slots}
\end{table*}

\paragraph{Context description.}
Each context caption $c^{\mathrm{ctx}}$ is obtained by filling age, sex, height, weight, BMI category, and acquisition time into one of several short sentence templates.
Representative abstract templates include:
\begin{enumerate}
    \item \texttt{\{AgeDesc\} \{SexTitle\} with \{BMI\} figure, \{Height\}, \{Weight\}, with data collected at \{Time\}.}
    \item \texttt{Data collected at \{Time\} shows \{Pronoun\} is \{AgeDesc\}, \{Height\}, \{Weight\}, and \{BMI\}.}
    \item \texttt{\{SexTitle\} individual recorded at \{Time\}, \{Age\} years old, \{Height\}, \{Weight\}, \{BMI\}.}
    \item \texttt{Recorded at \{Time\}: \{AgeDesc\}, \{SexTitle\}, \{Height\}, \{Weight\} with \{BMI\} shape.}
\end{enumerate}
Categorical labels follow Table~\ref{tab:appendix:rule}.

\paragraph{Instantiated examples.}
Below we fix one morphology feature set
\[
\begin{aligned}
&(T{=}0.96, \mathrm{IHR}{=}62.76, \mathrm{prop\_s}{=}0.16, \\
&\Delta T{=}0.32, t_{\mathrm{sys}}{=}0.38, t_{\mathrm{dia}}{=}0.58, \mathrm{AW}_s{=}1.20, \mathrm{AW}_d{=}0.80)
\end{aligned}
\]
and one metadata tuple
\[(\mathrm{age}{=}36, \mathrm{sex}{=}\mathrm{female}, \mathrm{height}{=}158, \mathrm{weight}{=}52, \mathrm{time}{=}08{:}19)\],
and show template-induced surface variation.

\begin{enumerate}
    \item \textit{Morph / order A:}
    Pulse characteristics: cardiac cycle 0.96s, 62.76 beats per minute within normal range, systolic upslope proportion 0.16, 0.32s duration systolic-diastolic peak interval, ventricular contraction 0.38s, relaxation 0.58s.
    \item \textit{Morph / order B:}
    PPG signal shows systolic upslope proportion 0.16, relaxation 0.58s, ventricular contraction 0.38s, cardiac cycle 0.96s, 62.76 bpm within normal range, systolic-diastolic peak interval measuring 0.32s.
    \item \textit{Morph / paraphrase:}
    Features observed in PPG: 0.96s pulse duration, instantaneous heart rate 62.76 bpm within normal range, proportion of systolic segment 0.16, delay between S and D waves measuring 0.32s, contraction phase 0.38s, diastolic period 0.58s.
    \item \textit{Morph / with AW:}
    Cardiovascular PPG traits include pulse cycle 0.96s, instantaneous heart rate 62.76 bpm within normal range, amplitude to width ratio: systolic 1.20, diastolic 0.80, systolic upslope proportion 0.16, time between systolic and diastolic peaks 0.32s.
    \item \textit{Context / age-first:}
    A 36-year-old young woman with a normal weight figure, with a height of 158 cm, 52 kg in weight, with data collected at morning 8:19.
    \item \textit{Context / time-first:}
    Data collected at morning 8:19 shows she is a 36-year-old young adult, with a height of 158 cm, 52 kg in weight, and normal weight.
    \item \textit{Context / title-first:}
    Woman individual recorded at morning 8:19, 36 years old, with a height of 158 cm, 52 kg in weight, normal weight.
    \item \textit{Paired sample:}
    $c^{\mathrm{morph}}$: Pulse characteristics: cardiac cycle 0.96s, 62.76 beats per minute within normal range, systolic upslope proportion 0.16, 0.32s duration systolic-diastolic peak interval, ventricular contraction 0.38s, relaxation 0.58s.
    $c^{\mathrm{ctx}}$: A 36-year-old young woman with a normal weight figure, with a height of 158 cm, 52 kg in weight, with data collected at morning 8:19.
\end{enumerate}

Table~\ref{tab:appendix:rule} summarizes the categorical mappings shared by context-description generation, Cross-Recording Context Alignment, and Zero-Shot User Profiling.
BMI follows Chinese adult cutoffs.

\begin{table*}[t]
    \centering
    \begin{tabular*}{0.7\textwidth}{@{\extracolsep{\fill}}ccc@{}}
        \toprule
        Variable & Condition & Category \\
        \midrule
        \multirow{4}{*}{Age}
        & Age $< 18$ & adolescent \\
        & $18 \le \mathrm{Age} < 30$ & young adult \\
        & $30 \le \mathrm{Age} < 50$ & middle-aged \\
        & Age $\ge 50$ & elderly \\
        \midrule
        Sex & male / female & male / female \\
        \midrule
        \multirow{4}{*}{BMI}
        & BMI $< 18.5$ & underweight \\
        & $18.5 \le \mathrm{BMI} < 24$ & normal weight \\
        & $24 \le \mathrm{BMI} < 28$ & overweight \\
        & BMI $\ge 28$ & obese \\
        \midrule
        \multirow{7}{*}{Hour of day}
        & Hour $\ge 22$ or Hour $< 4$ & nighttime \\
        & $4 \le \mathrm{Hour} < 7$ & sunrise dawn \\
        & $7 \le \mathrm{Hour} < 11$ & morning \\
        & $11 \le \mathrm{Hour} \le 14$ & midday \\
        & $15 \le \mathrm{Hour} < 18$ & afternoon \\
        & $18 \le \mathrm{Hour} < 20$ & sunset dusk \\
        & $20 \le \mathrm{Hour} < 22$ & evening \\
        \bottomrule
        \multicolumn{3}{@{}l@{}}{\footnotesize BMI $= \mathrm{weight}/(\mathrm{height}/100)^2$ with height in cm and weight in kg.} \\
    \end{tabular*}
    \caption{Discretization rules for metadata categories.}
    \label{tab:appendix:rule}
\end{table*}

\section{More Details about \method}
\label{sec:appendix:method_details}

\subsection{Metadata Grouping and Aggregation}
\label{sec:appendix:grouping}

Each recording is assigned a four-field categorical metadata key
\begin{equation}
    a_i = \left(a_i^{\mathrm{age}}, a_i^{\mathrm{sex}},
    a_i^{\mathrm{BMI}}, a_i^{\mathrm{time}}\right),
\end{equation}
where each field is represented by its discretized category index.
Within each minibatch $\mathcal{B}$, samples are partitioned using their metadata keys:
\begin{equation}
    \mathcal{G}_{a} = \{i \in \mathcal{B} \mid a_i = a\}.
\end{equation}
In implementation, the four categorical indices are serialized as a dictionary key, and the PPG embeddings associated with the same key are stacked before aggregation.

For each observed tuple $a$, a single learnable query $q$ produces one scalar score for every recording embedding in the group.
The group representation is
\begin{align}
    \alpha_i & =
    \frac{\exp\left(q \top h_i^p\right)}
    {\sum_{j\in\mathcal{G}_{a}}\exp\left(q \top h_j^p\right)}, \\
    h_a^p & = \sum_{i\in\mathcal{G}_{a}} \alpha_i h_i^p.
\end{align}
The aggregated PPG representation is aligned with a context description sampled from the same metadata-tuple group.

\subsection{Attribute-Aware Masking for WGTM}
\label{sec:appendix:wgtm_masking}

Waveform-Guided Text Modeling masks complete attribute-specific spans rather than independently masking arbitrary tokens.
Because the context descriptions are generated from deterministic templates, we construct separate regular expressions for age, sex, BMI, and acquisition time.
For a targeted attribute, every matched span is replaced by \texttt{[MASK]} tokens, including both its categorical descriptor and any associated numerical value.
For example, age masking removes both the numerical age and the age-group phrase.
When BMI is targeted, we additionally mask the height and weight values and their corresponding descriptive spans, since these fields can otherwise reveal the BMI category without using the PPG input.

The WGTM objective is evaluated only at masked token positions and unmasked tokens are excluded from the loss.
This masking strategy removes direct textual shortcuts and requires the decoder to recover the selected contextual concepts from the remaining text together with the PPG patch representations.

\section{Model Architecture and Training Hyperparameters}
\label{sec:appendix:implementation}



\subsection{PPG Encoder}

The PPG encoder is a 12-layer Transformer with hidden dimension 512, eight attention heads, an MLP expansion ratio of 4, and dropout 0.1.
The input waveform is divided into non-overlapping patches using a patch size and stride of 125, corresponding to 0.5 seconds per patch and 24 patch tokens per input.
A absolute positional embeddings are added before the Transformer blocks.
The final \texttt{[CLS]} representation is used as the recording-level embedding.
An MLP projection head maps this representation into the shared PPG-Text embedding space.

\subsection{Text Encoder and Decoder}

We initialize the text model from the 12-layer PubMedBERT architecture.
The bottom six Transformer layers form the text encoder.
The top six layers form the text decoder, with an additional cross-attention module inserted into each layer to condition text reconstruction on PPG patch representations.
Before serving as keys and values in cross-attention, PPG patch features are mapped to the same dimension as the text hidden space by a trainable linear projection.
An MLP projection head maps the encoded text representation into the shared embedding space.

The pretrained PubMedBERT parameters are frozen by default.
We optimize all LayerNorm parameters, the final layer of the six-layer text encoder, the newly inserted decoder cross-attention modules, and the PPG-to-text linear projection.
The pretrained self-attention and feed-forward parameters in the decoder remain frozen.

\subsection{Pretraining Configuration}

We pretrain \method for 100 epochs with a minibatch size of 512 on four NVIDIA A100-80GB GPUs using \texttt{bfloat16} arithmetic.
Optimization uses AdamW with learning rate $1\times10^{-4}$, weight decay 0.1, and $\beta_1=0.9$, $\beta_2=0.999$.
The learning rate is linearly warmed up for 1{,}000 steps and then decayed with a cosine schedule.
The learnable contrastive scaling parameter is initialized to 2.659.
All three objectives use equal weights.

\section{Downstream Evaluation}
\label{sec:appendix:downstream}

\begin{table*}[t]
    \centering
    \small
    \setlength{\tabcolsep}{4pt}
    \begin{tabular}{l|ccccc}
        \toprule
        Method
        & SBP MAE$\downarrow$
        & DBP MAE$\downarrow$
        & cfPWV PR-AUC$\uparrow$
        & Age MAE$\downarrow$
        & Sex ROC-AUC$\uparrow$ \\
        \midrule
        SimCLR~\cite{simclr}            & 12.40$\pm$0.77 & \underline{6.92$\pm$0.29} & 0.70$\pm$0.23 & 8.02$\pm$0.79 & 0.76$\pm$0.09 \\
        MoCo-v3~\cite{mocov3}           & 12.75$\pm$1.10 & 7.19$\pm$0.43 & 0.71$\pm$0.21 & 9.28$\pm$1.30 & 0.75$\pm$0.09 \\
        MAE~\cite{mae}                  & 12.61$\pm$0.47 & 7.21$\pm$0.19 & \textbf{0.76$\pm$0.26} & \underline{7.15$\pm$0.82} & 0.79$\pm$0.05 \\
        PaPaGei~\cite{papagei}          & 12.62$\pm$0.53 & 7.34$\pm$0.22 & 0.73$\pm$0.24 & 7.66$\pm$0.88 & \underline{0.80$\pm$0.07} \\
        Apple~\cite{apple0}             & \textbf{12.24$\pm$0.89} & 6.98$\pm$0.30 & 0.73$\pm$0.26 & 8.54$\pm$1.42 & 0.77$\pm$0.08 \\
        \midrule
        SupCon~\cite{khosla2020supcon}  & 12.27$\pm$0.52 & 7.07$\pm$0.22 & 0.64$\pm$0.23 & 8.23$\pm$1.00 & \textbf{0.82$\pm$0.08} \\
        Metadata Embedding              & 13.82$\pm$0.61 & 7.74$\pm$0.29 & 0.64$\pm$0.25 & 7.98$\pm$0.60 & 0.79$\pm$0.10 \\
        Multi-Task Supervision          & 14.33$\pm$0.48 & 8.34$\pm$0.14 & \underline{0.74$\pm$0.17} & 7.46$\pm$0.73 & 0.79$\pm$0.06 \\
        \midrule
        Morphology Alignment            & 13.16$\pm$0.65 & 7.42$\pm$0.20 & 0.72$\pm$0.19 & 8.04$\pm$0.95 & 0.74$\pm$0.04 \\
        + Context Alignment             & 12.84$\pm$0.52 & 7.07$\pm$0.18 & 0.72$\pm$0.21 & 7.19$\pm$0.84 & \underline{0.80$\pm$0.08} \\
        + WGTM (\method)                & \underline{12.26$\pm$0.48} & \textbf{6.88$\pm$0.17} & 0.67$\pm$0.25 & \textbf{7.14$\pm$0.90} & 0.79$\pm$0.08 \\
        \bottomrule
    \end{tabular}
    \caption{Linear-probing results with 10\% of the labeled training data. Results follow the same evaluation protocol as the full-data experiments. Best results are in \textbf{bold}; second-best results are \underline{underlined}. Ties are marked equally.}
    \label{tab:appendix:few_shot_10}
\end{table*}

\begin{table*}[t]
    \centering
    \small
    \setlength{\tabcolsep}{4pt}
    \begin{tabular}{l|ccccc}
        \toprule
        Method
        & SBP MAE$\downarrow$
        & DBP MAE$\downarrow$
        & cfPWV PR-AUC$\uparrow$
        & Age MAE$\downarrow$
        & Sex ROC-AUC$\uparrow$ \\
        \midrule
        SimCLR~\cite{simclr}            & 12.14$\pm$0.84 & 6.75$\pm$0.30 & 0.68$\pm$0.12 & 7.91$\pm$0.80 & 0.77$\pm$0.07 \\
        MoCo-v3~\cite{mocov3}           & 12.24$\pm$0.83 & 6.79$\pm$0.29 & 0.66$\pm$0.08 & 8.75$\pm$1.14 & 0.76$\pm$0.08 \\
        MAE~\cite{mae}                  & 12.15$\pm$0.32 & 6.94$\pm$0.16 & \underline{0.77$\pm$0.06} & \underline{7.14$\pm$0.76} & 0.81$\pm$0.08 \\
        PaPaGei~\cite{papagei}          & 12.89$\pm$0.48 & 7.17$\pm$0.19 & 0.76$\pm$0.10 & 7.70$\pm$0.84 & 0.80$\pm$0.07 \\
        Apple~\cite{apple0}             & \underline{11.97$\pm$0.80} & \textbf{6.62$\pm$0.20} & 0.74$\pm$0.17 & 7.63$\pm$0.78 & 0.78$\pm$0.09 \\
        \midrule
        SupCon~\cite{khosla2020supcon}  & 12.12$\pm$0.53 & 6.96$\pm$0.15 & 0.73$\pm$0.10 & 7.81$\pm$0.76 & \underline{0.82$\pm$0.08} \\
        Metadata Embedding              & 12.47$\pm$0.60 & 7.03$\pm$0.20 & 0.68$\pm$0.17 & 7.72$\pm$0.64 & 0.81$\pm$0.09 \\
        Multi-Task Supervision          & 13.26$\pm$0.41 & 7.64$\pm$0.16 & \textbf{0.78$\pm$0.11} & 7.36$\pm$0.71 & 0.81$\pm$0.07 \\
        \midrule
        Morphology Alignment            & 12.21$\pm$0.55 & 6.92$\pm$0.17 & 0.72$\pm$0.15 & 7.96$\pm$1.08 & 0.78$\pm$0.06 \\
        + Context Alignment             & 12.00$\pm$0.48 & 6.81$\pm$0.13 & 0.76$\pm$0.12 & 7.15$\pm$0.88 & 0.81$\pm$0.07 \\
        + WGTM (\method)                & \textbf{11.86$\pm$0.51} & \underline{6.73$\pm$0.20} & \textbf{0.78$\pm$0.09} & \textbf{7.02$\pm$0.85} & \textbf{0.83$\pm$0.06} \\
        \bottomrule
    \end{tabular}
    \caption{Linear-probing results with 30\% of the labeled training data. Results follow the same evaluation protocol as the full-data experiments. Best results are in \textbf{bold}; second-best results are \underline{underlined}. Ties are marked equally.}
    \label{tab:appendix:few_shot_30}
\end{table*}

\subsection{Linear Probing}
\label{sec:appendix:linear_probe}

For all linear-probing experiments, the pretrained PPG encoder is frozen and only a task-specific linear prediction head is optimized.
Unless otherwise stated, the downstream optimizer is AdamW with learning rate $1\times10^{-3}$, weight decay $1\times10^{-4}$, batch size 256, and a maximum of 50 epochs.
Model selection uses only the validation set.

\paragraph{Blood pressure, age, and sex on HBPM-X.}
We use five-fold cross-validation for SBP regression, DBP regression, age regression, and sex classification.
SBP and DBP use separate linear outputs, and their mean-squared-error losses are equally weighted.
Age regression also uses mean-squared error, while sex classification uses binary cross-entropy with logits.
Training is stopped when the validation loss does not improve for five epochs, and the checkpoint with the lowest validation loss is retained.
We report MAE and $R^2$ for regression, and accuracy and ROC-AUC for sex classification.
Blood pressure is evaluated at the recording level.
For age and sex, predictions from all recordings belonging to the same subject are averaged before computing subject-level metrics.
The final result is the mean and standard deviation over the five folds.

\paragraph{Arterial-stiffness classification on cfPWV.}
We construct five independent subject-level stratified splits using a 7:1:2 train/validation/test ratio.
Each split contains 392 training subjects (296 normal and 96 abnormal), 56 validation subjects (42 normal and 14 abnormal), and 112 test subjects (85 normal and 27 abnormal).
All recordings from one subject remain in the same split.
The linear classifier is trained with class-weighted binary cross-entropy, where the positive-class weight is the negative-to-positive subject ratio in the training set.
Early stopping and checkpoint selection are based on validation subject-level binary cross-entropy, with patience 5 and minimum improvement $10^{-3}$.

At inference, record-level logits from the same subject are averaged to obtain one subject-level score.
For threshold-dependent metrics, the decision threshold is selected on the validation subjects by maximizing Youden's $J$ statistic and is then applied unchanged to the test set.
We report balanced accuracy and PR-AUC, summarized as mean and standard deviation over the five splits.

\paragraph{Heart-rate regression on PPG-DaLiA.}
We follow leave-one-subject-out evaluation over the 15 subjects.
In each fold, one subject is held out for testing, and another one is deterministically assigned to validation and the other 13 are used for training.
No subject is divided across training, validation, and test sets.
The linear regressor is trained with mean-squared error, and the checkpoint with the lowest validation MSE is selected.
We report test MAE and $R^2$.

Each fold is repeated with multiple random seeds, and the test metrics are first averaged across seeds within that fold.
The final score is reported as mean $\pm$ standard deviation over the 15 held-out subjects.
Consequently, the reported standard deviation primarily reflects cross-subject variability under the leave-one-subject-out protocol.

\subsection{Zero-Shot User Profiling}
\label{sec:appendix:zero_shot}

Zero-shot user profiling is evaluated on HBPM-X without fitting a downstream predictor.
We enumerate the Cartesian product of four age groups, two sex categories, four BMI categories, and seven acquisition-time categories.
Each tuple is instantiated using the context-description template and encoded by the text encoder.

For each subject, multiple PPG recordings are encoded and aggregated with the same learned single-query attention aggregator used during pretraining.
Unless otherwise stated, the main profiling result uses 10 recordings per subject.
We retrieve the three candidate texts with the highest similarity to the aggregated PPG representation.
The age group, sex, BMI category, and acquisition-time category are predicted independently by majority voting over the corresponding fields of the top-three candidates.
Acquisition time participates in candidate construction and retrieval but is marginalized in the final joint metric.
Specifically, joint accuracy counts a prediction as correct only when age group, sex, and BMI category are all correct, irrespective of the predicted acquisition-time category.

\section{Additional Experiments}
\label{sec:appendix:additional_experiments}

\subsection{Low-Resource Linear Probing}
\label{sec:appendix:few_shot}

We further evaluate label efficiency under 10\% and 30\% labeled-data regimes. For each downstream task, we retain only the corresponding fraction of the labeled training partition while keeping the validation and test partitions, frozen encoder, prediction head, optimization settings, and model-selection protocol unchanged from Sec.~\ref{sec:appendix:linear_probe}.

Table~\ref{tab:appendix:few_shot_10} shows that, with only 10\% labeled data, \method obtains the lowest DBP and age MAE and the second-lowest SBP MAE. Performance on cfPWV is more variable in this regime, consistent with the limited number of labeled subjects and the class imbalance of the cohort. With 30\% labeled data, Table~\ref{tab:appendix:few_shot_30} shows that \method achieves the best SBP, age, and sex results, ties for the best cfPWV PR-AUC, and ranks second on DBP. Within the cumulative variants, Context Alignment and WGTM improve all five metrics over Morphology Alignment at the 30\% label fraction. These results indicate that the pretrained representation remains useful when downstream annotation is limited.

\subsection{Effect of Pretraining Data Scale}
\begin{figure}[th]
    \centering
    \includegraphics[width=\linewidth]{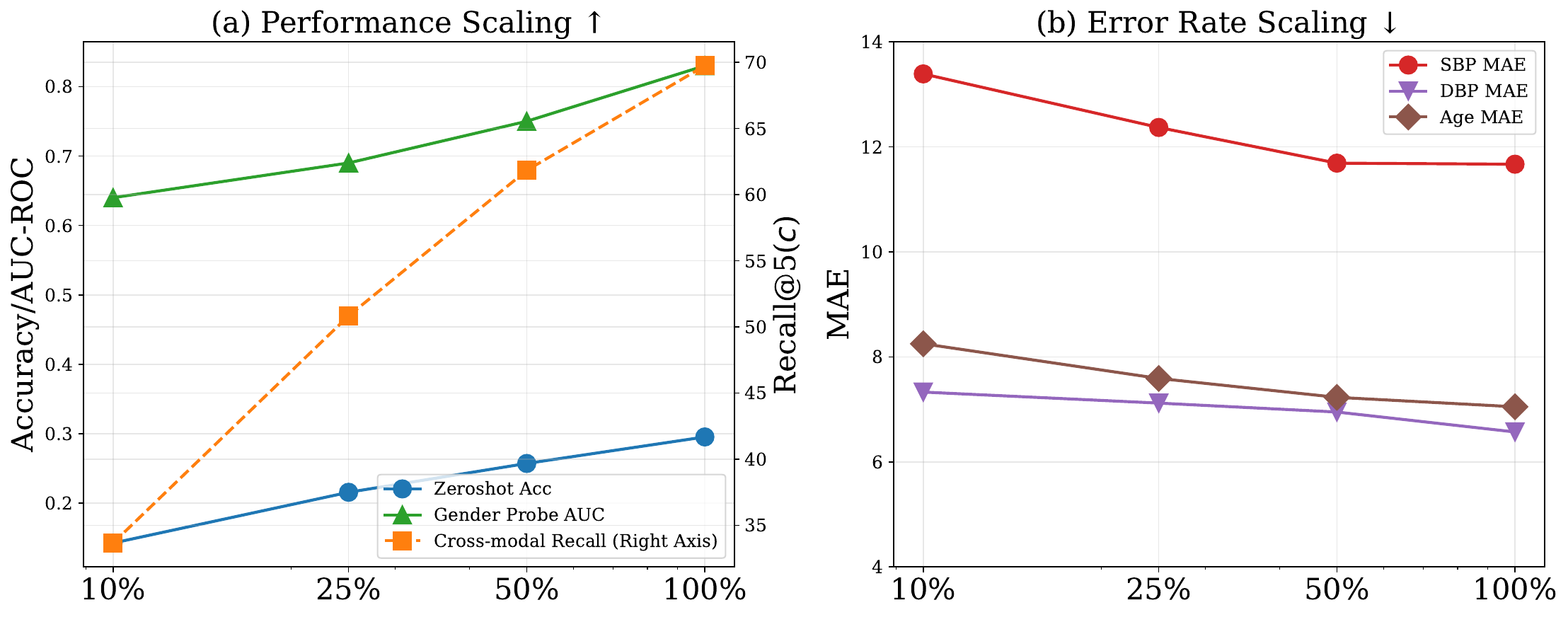}
    \caption{Effect of pretraining data scale. \method is pretrained using 10\%, 25\%, 50\%, and 100\% of the full corpus. (a) Zero-shot accuracy, recorded-sex probe AUROC, and cross-modal Recall@5. (b) SBP, DBP, and age MAE. Higher is better in (a), whereas lower is better in (b).}
    \label{fig:exp:scaling}
\end{figure}

We examine how \method benefits from additional pretraining data by training on 10\%, 25\%, 50\%, and 100\% subsets of the full corpus. As shown in Fig.~\ref{fig:exp:scaling}, increasing the data fraction generally improves zero-shot profiling, recorded-sex probing, and cross-modal retrieval, while reducing the MAE of SBP, DBP, and age prediction. The largest relative improvement is observed in cross-modal retrieval, suggesting that learning a well-structured PPG–language space is particularly data dependent. These results indicate that \method continues to benefit from broader pretraining coverage, without assuming a specific scaling-law functional form.

\subsection{Leave-One-Attribute-Out Pretraining}
\label{sec:appendix:loao}

We examine whether downstream transfer is dominated by the literal occurrence of target attributes in the pretraining descriptions. We train three leave-one-attribute-out variants. For \method w/o Age and \method w/o Sex, the corresponding numerical values and categorical phrases are removed from context descriptions and are excluded from WGTM targets. The metadata tuple used for Cross-Recording Context Alignment is left unchanged so that the grouping structure and training batches remain comparable. These two variants therefore isolate direct textual exposure. For \method w/o HR, the explicit instantaneous-heart-rate field and its qualitative descriptor are removed from the morphology descriptions.

Tables~\ref{tab:appendix:loao_bp_pwv} and~\ref{tab:appendix:loao_age_sex_hr} show that removing one explicit attribute does not cause the corresponding downstream task to collapse. Age MAE changes from $7.05$ to $7.22$ without age text, sex ROC-AUC changes from $0.83$ to $0.82$ when leaving sex out, and external HR MAE changes from $7.65$ to $7.70$ with heart-rate field removed. Thus, the transfer results are not solely a direct readout of the corresponding textual field.

The ablations nevertheless affect tasks beyond the removed attribute. In particular, omitting age produces the largest degradation, with SBP MAE increases from $11.67$ to $11.88$, DBP MAE from $6.57$ to $6.70$, age MAE from $7.05$ to $7.22$, and PPG-DaLiA HR MAE from $7.65$ to $8.55$. Removing sex has a smaller overall effect. Removing the explicit HR field also degrades DBP and cfPWV performance despite leaving external HR transfer nearly unchanged. These results suggest that individual attributes act as complementary organizing cues for the learned physiological space, with age providing particularly broad contextual information, rather than serving only as isolated prediction targets.

\subsection{Lights-Out Diagnosis of WGTM}
\label{sec:appendix:wgtm_lightsout}

We test whether masked-token recovery depends on informative PPG features or can be performed from the residual text alone. The pretrained checkpoint is kept fixed, and only the PPG encoder states supplied as keys and values to decoder cross-attention are modified. We compare \textbf{Normal}, which uses the actual PPG hidden states; \textbf{Zero}, which replaces all states with zeros; and \textbf{Random}, which replaces them with Gaussian noise of the same shape. Accuracy is computed only at the attribute-aware masked positions used by the WGTM objective.

\begin{table}[!h]
    \centering
    \small
    \begin{tabular}{lc}
        \toprule
        PPG encoder states & Masked-token Acc.$\uparrow$ \\
        \midrule
        Normal & 0.709 \\
        Zero & 0.172 \\
        Random & 0.186 \\
        \midrule
        Drop (Normal $-$ Zero) & 0.537 \\
        \bottomrule
    \end{tabular}
    \caption{Lights-out diagnosis of WGTM under different PPG encoder states supplied to decoder cross-attention.}
    \label{tab:appendix:wgtm_lightsout}
\end{table}
As shown in Table~\ref{tab:appendix:wgtm_lightsout}, masked-token accuracy is $0.709$ with the actual PPG states, but falls to $0.172$ with zero states and $0.186$ with random states. The $0.537$ absolute drop from Normal to Zero shows that the unmasked textual context alone is insufficient for most recoveries. The similarly low Zero and Random results further indicate that non-informative cross-attention inputs do not provide a useful fallback. This diagnostic supports the intended role of WGTM, where the recovery of masked contextual concepts encourages finer-grained PPG representation learning.


\begin{table*}[t]
    \centering
    \small
    \setlength{\tabcolsep}{4pt}
    \begin{tabular}{l|cc|cc|cc}
        \toprule
        \multirow{2}{*}{Method}
        & \multicolumn{2}{c|}{SBP}
        & \multicolumn{2}{c|}{DBP}
        & \multicolumn{2}{c}{cfPWV} \\
        & MAE$\downarrow$ & $R^2\uparrow$
        & MAE$\downarrow$ & $R^2\uparrow$
        & Balanced ACC$\uparrow$ & PR-AUC$\uparrow$ \\\midrule
        \method                    & 11.67$\pm$0.48 & 0.11$\pm$0.03 & 6.57$\pm$0.11 & 0.11$\pm$0.05 & 0.85$\pm$0.02 & 0.84$\pm$0.03 \\
        \method w/o Age            & 11.88$\pm$0.50 & 0.08$\pm$0.04 & 6.70$\pm$0.16 & 0.08$\pm$0.05 & 0.84$\pm$0.03 & 0.80$\pm$0.03 \\
        \method w/o Sex            & 11.79$\pm$0.43 & 0.10$\pm$0.03 & 6.65$\pm$0.14 & 0.09$\pm$0.03 & 0.85$\pm$0.03 & 0.81$\pm$0.04 \\
        \method w/o HR             & 11.77$\pm$0.46 & 0.11$\pm$0.04 & 6.73$\pm$0.16 & 0.11$\pm$0.03 & 0.84$\pm$0.04 & 0.79$\pm$0.03 \\
        \bottomrule
    \end{tabular}
    \caption{Linear-probing transfer after removing one explicit attribute from pretraining descriptions. The age- and sex-related texts are removed from context description.}
    \label{tab:appendix:loao_bp_pwv}
\end{table*}

\begin{table*}[t]
    \centering
    \small
    \setlength{\tabcolsep}{4pt}
    \begin{tabular}{l|cc|cc|cc}
        \toprule
        \multirow{2}{*}{Method}
        & \multicolumn{2}{c|}{Age}
        & \multicolumn{2}{c|}{Sex}
        & \multicolumn{2}{c}{HR} \\
        & MAE$\downarrow$ & $R^2\uparrow$
        & ACC$\uparrow$ & ROC-AUC$\uparrow$
        & MAE$\downarrow$ & $R^2\uparrow$ \\
        \midrule
        \method                    & 7.05$\pm$0.79 & 0.42$\pm$0.11 & 0.74$\pm$0.06 & 0.83$\pm$0.08 & 7.65$\pm$3.09 & 0.49$\pm$0.36 \\
        \method w/o Age            & 7.22$\pm$0.81 & 0.40$\pm$0.11 & 0.71$\pm$0.07 & 0.82$\pm$0.07 & 8.55$\pm$3.84 & 0.38$\pm$0.54 \\
        \method w/o Sex            & 7.08$\pm$0.87 & 0.42$\pm$0.11 & 0.74$\pm$0.06 & 0.82$\pm$0.09 & 7.72$\pm$3.60 & 0.47$\pm$0.51 \\
        \method w/o HR             & 7.11$\pm$0.82 & 0.42$\pm$0.10 & 0.74$\pm$0.06 & 0.83$\pm$0.08 & 7.70$\pm$3.59 & 0.48$\pm$0.47 \\
        \bottomrule
    \end{tabular}
    \caption{Attribute and external heart-rate probing after leave-one-attribute-out pretraining. The HR ablation removes the explicit instantaneous-heart-rate field from morphology descriptions.}
    \label{tab:appendix:loao_age_sex_hr}
\end{table*}

\begin{table}[!b]
    \centering
    \scriptsize
    \setlength{\tabcolsep}{1pt}
    \begin{tabular*}{\columnwidth}{@{\extracolsep{\fill}}lccccc@{}}
        \toprule
        Aggregation & $N$=1 & $N$=2 & $N$=5 & $N$=10 & $N$=15 \\
        \midrule
        Attention & 5.10$\pm$0.42 & \textbf{10.21$\pm$1.09} & \textbf{21.40$\pm$2.37} & \textbf{29.55$\pm$0.80} & \textbf{29.47$\pm$1.89} \\
        Mean      & \textbf{5.57$\pm$1.60} & 9.31$\pm$1.96 & 17.46$\pm$1.81 & 25.81$\pm$2.21 & 27.49$\pm$2.18 \\
        Max       & 4.88$\pm$0.72 & 8.60$\pm$1.36 & 14.40$\pm$2.15 & 12.84$\pm$1.53 & 13.00$\pm$1.96 \\
        \bottomrule
    \end{tabular*}
    \caption{Joint zero-shot profiling accuracy (\%) for different aggregation operators and numbers of PPG recordings per subject. Joint correctness requires age group, sex, and BMI category to be simultaneously correct, where acquisition time is marginalized. All results use top-$3$ candidate retrieval. The best mean result in each column is shown in \textbf{bold}.}
    \label{tab:appendix:aggregation_ablation}
\end{table}

\subsection{Top-$K$ Robustness of Zero-Shot Profiling}
\label{sec:appendix:topk_robustness}

We examine whether zero-shot profiling depends sensitively on the number of retrieved context descriptions. Following the protocol in Sec.~\ref{sec:appendix:zero_shot}, we fix 10 PPG recordings per subject and vary the retrieval depth over $k\in\{1,\ldots,7\}$. Age group, sex, and BMI category are predicted by field-wise majority voting over the top-$k$ candidates, and joint accuracy requires all three attributes to be correct.

\begin{figure}[t]
    \centering
    \includegraphics[width=0.65\linewidth]{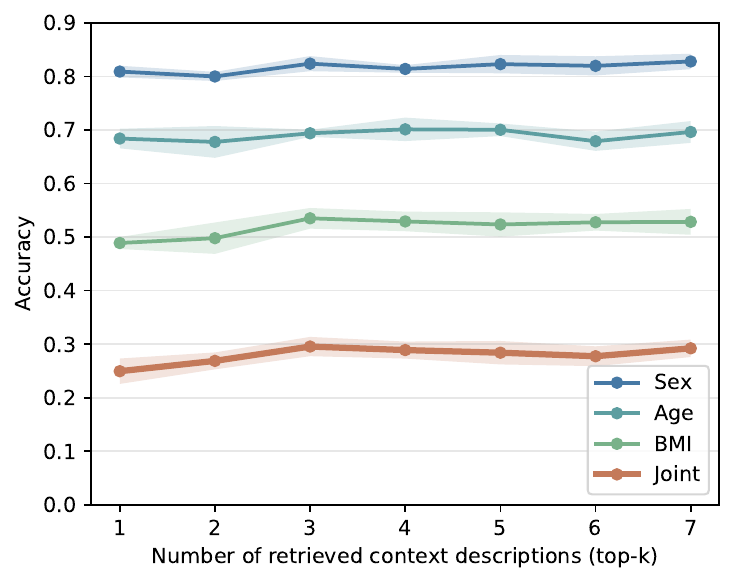}
    \caption{Zero-shot profiling accuracy versus retrieval depth $k$ with 10 PPG recordings per subject. Shaded regions denote one standard deviation over five random trials.}
    \label{fig:appendix:topk}
\end{figure}

As shown in Fig.~\ref{fig:appendix:topk}, performance remains stable across the full range of $k$. Sex accuracy stays near $0.81$--$0.83$ and age accuracy near $0.68$--$0.70$. BMI and joint accuracy improve modestly from $k=1$ to $k=3$ (joint: $24.9\%$ to $29.6\%$) and then plateau, without degradation as additional lower-ranked candidates enter the vote. These trends indicate that the zero-shot protocol is robust to the choice of top-$k$, with $k=3$ providing a favorable operating point used throughout the main evaluation.

\subsection{Cross-Recording Aggregation}
\label{sec:appendix:aggregation_ablation}

We further separate the effect of accumulating multiple recordings from the choice of subject-level aggregation operator in zero-shot profiling. Under the same top-$3$ retrieval protocol described in Sec.~\ref{sec:appendix:zero_shot}, we vary the number of PPG recordings per subject as $N\in\{1,2,5,10,15\}$. The learned single-query attention aggregator is compared with mean pooling and element-wise max pooling. Mean and max pooling are inference-time replacements applied to the same pretrained representation model, and no additional training is performed. We report joint accuracy for age group, sex, and BMI category, while acquisition time remains marginalized as in the main evaluation.

Table~\ref{tab:appendix:aggregation_ablation} shows that both attention and mean pooling benefit substantially from additional recordings. Attention aggregation increases from $5.10\%$ with one recording to $21.40\%$ with five and $29.55\%$ with ten, after which it saturates. Mean pooling follows the same overall trend, reaching $27.49\%$ with 15 recordings, but remains below attention pooling for $N\geq 2$ except within the variation at the largest setting. The agreement between these two aggregation rules indicates that the principal gain comes from accumulating repeated waveform evidence rather than from a single specialized pooling operator. The learned attention query provides an additional advantage by weighting recordings according to their contribution to the shared representation.

In contrast, max pooling improves only up to five recordings and then decreases to approximately $13\%$. Element-wise maxima can preserve record-specific extremes instead of reinforcing recurring structure, making it less suitable for suppressing transient variation across recordings. This results also support the evidence-scope motivation of \method: contextual semantics become more reliably accessible as multiple observations are integrated, where a learned or averaging-based aggregator is preferable to obtain a contextual reliable representation.

\section{Failure Mode Analysis}
\label{sec:appendix:failure_modes}

Hard discretization of continuous physiology can create brittle decision boundaries even when overall transfer metrics remain strong. We examine two such cases: clinical thresholding on cfPWV and categorical BMI prediction in zero-shot profiling.

\subsection{Errors Near the cfPWV Clinical Cutoff}
\label{sec:appendix:failure_cfpwv}

Arterial-stiffness labels are obtained by thresholding continuous cfPWV at $10\,\mathrm{m/s}$, so subjects near this cutoff are intrinsically ambiguous. Using the same subject-level protocol and validation-tuned decision thresholds as the main evaluation, we compare test error for subjects with $|\mathrm{cfPWV}-10|<2$ against those farther from the cutoff.

\begin{figure}[t]
    \centering
    \includegraphics[width=0.55\linewidth]{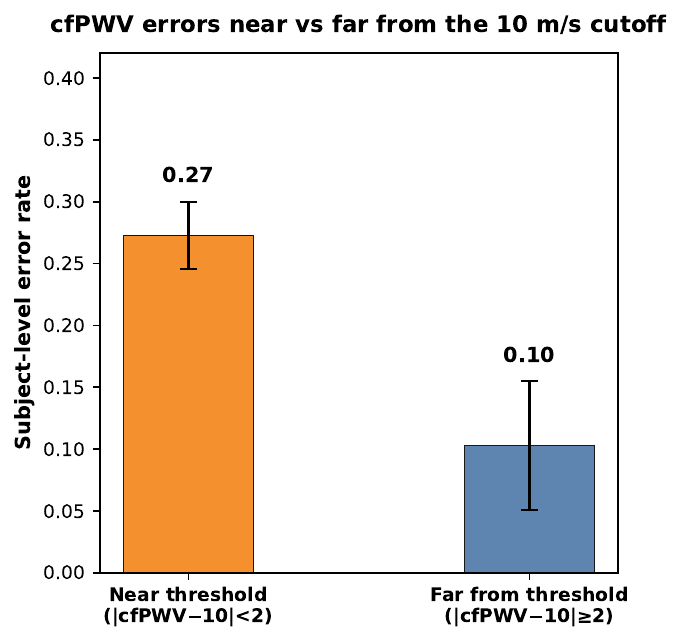}
    \caption{Subject-level cfPWV classification error near versus far from the $10\,\mathrm{m/s}$ clinical cutoff. Bars show mean $\pm$ std over five splits (on average 54 near-threshold and 58 far-from-threshold test subjects per split).}
    \label{fig:appendix:failure_cfpwv}
\end{figure}

As shown in Fig.~\ref{fig:appendix:failure_cfpwv}, the near-threshold error rate is $0.27\pm0.03$, about $2.7\times$ higher than the far-from-threshold rate of $0.10\pm0.05$. The near-threshold error coincides with the region where continuous cfPWV values afford the least decisive evidence for a binary label. In this sense, the result characterizes residual uncertainty around the clinical boundary under hard thresholding.

\subsection{BMI Confusion in Zero-Shot Profiling}
\label{sec:appendix:failure_bmi}

\begin{figure}[t]
    \centering
    \includegraphics[width=0.55\linewidth]{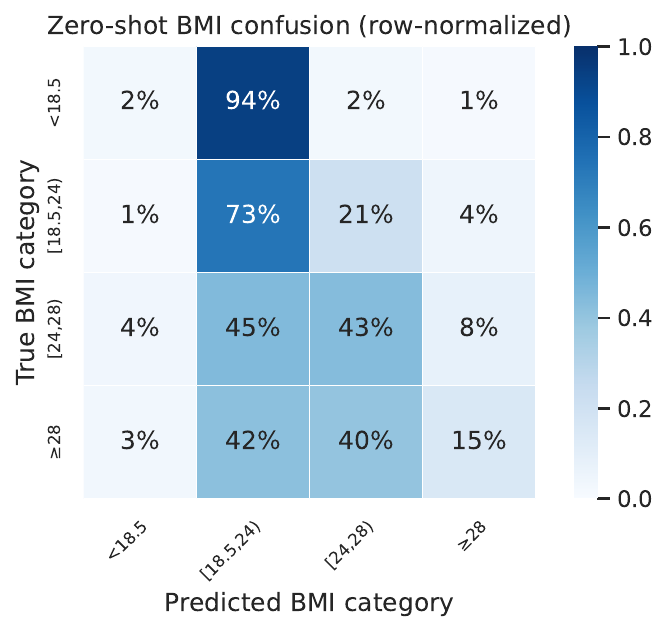}
    \caption{Row-normalized BMI confusion matrix for zero-shot profiling, averaged over five random trials.}
    \label{fig:appendix:failure_bmi}
\end{figure}

Among zero-shot attributes, BMI is the weakest, with accuracy of approximately $0.52$, compared with about $0.81$ for sex and $0.70$ for age. Fig.~\ref{fig:appendix:failure_bmi} reports the row-normalized BMI confusion matrix under the main zero-shot protocol with $N{=}10$ recordings and top-$3$ retrieval.

Normal-weight subjects are recovered most often, whereas underweight and obese subjects are frequently assigned to more central categories. This pattern is consistent with a majority-class prior in the evaluation cohort and with the comparatively weak body-size cues available in wrist PPG. Discrete BMI categories remain useful as contextual descriptors during pretraining, but their zero-shot readout, especially for minority bins, should be interpreted cautiously as a coarse attribute cue rather than a precise body-shape assessment.

\bibliography{aaai2027}

\end{document}